\documentclass[10pt,journal,compsoc]{IEEEtran}
\ifCLASSOPTIONcompsoc
  \usepackage[nocompress]{cite}
\else
  \usepackage{cite}
\fi
\ifCLASSINFOpdf
  \usepackage[pdftex]{graphicx}
  \graphicspath{{./figs/}}
  \DeclareGraphicsExtensions{.pdf,.jpeg,.png}
\else
\fi
\usepackage[cmex10]{amsmath}
\usepackage{amsfonts}
\usepackage{algorithmic}
\usepackage{color}
\usepackage{array}
\ifCLASSOPTIONcompsoc
  \usepackage[caption=false,font=footnotesize,labelfont=sf,textfont=sf]{subfig}
\else
  \usepackage[caption=false,font=footnotesize]{subfig}
\fi
\usepackage{multirow}
\usepackage{footnote}
\makesavenoteenv{tabular}
\makesavenoteenv{table}
\begin{document}
\bstctlcite{IEEEexample:BSTcontrol}
\title{Measuring and Predicting Tag Importance for Image Retrieval}
\author{Shangwen~Li, 
        Sanjay~Purushotham, 
        Chen~Chen, 
        Yuzhuo~Ren,~\IEEEmembership{Student Member,~IEEE,}
        and~ C.-C.~Jay~Kuo,~\IEEEmembership{Fellow,~IEEE}
\IEEEcompsocitemizethanks{\IEEEcompsocthanksitem S. Li, S. Purushotham,  C. Chen, Y. Ren, and C. Kuo are with the Ming Hsieh Department of Electrical Engineering, University of Southern California, Los Angeles, CA 90089. E-mail: \{shangwel, yuzhuore\}@usc.edu, \{sanjayp2005, ohyline\}@gmail.com, cckuo@sipi.usc.edu\protect}
}


\IEEEtitleabstractindextext{%
\begin{abstract}
Textual data such as tags, sentence descriptions are combined with visual cues to reduce the semantic gap for image retrieval applications in today's
Multimodal Image Retrieval (MIR) systems. However, all tags are treated as equally important in
these systems, which may result in misalignment between visual and
textual modalities during MIR training. This will further lead to
degenerated retrieval performance at query time. To address this issue,
we investigate the problem of tag importance prediction, where the goal
is to automatically predict the tag importance and use it in image
retrieval. To achieve this, we first propose a method to measure
the relative importance of object and scene tags from image sentence
descriptions. Using this as
the ground truth, we present a tag importance prediction model to jointly exploit visual, semantic and context cues. The Structural
Support Vector Machine (SSVM) formulation is adopted to ensure efficient
training of the prediction model. Then, the Canonical Correlation
Analysis (CCA) is employed to learn the relation between the image
visual feature and tag importance to obtain robust retrieval performance. Experimental results on three real-world datasets show a significant
performance improvement of the proposed MIR with Tag Importance Prediction
(MIR/TIP) system over other MIR systems. 
\end{abstract}

}

\maketitle
\IEEEdisplaynontitleabstractindextext
\IEEEpeerreviewmaketitle

\IEEEraisesectionheading{\section{Introduction}\label{sec:introduction}}

\IEEEPARstart{I}{mage} retrieval \cite{rui1999image} is a long-standing
problem in the computer vision and information retrieval fields.
Current image search engines in service rely heavily on the text data.
It attempts to match user-input keywords (tags) to accompanying texts of
an image. It is thus called Tag Based Image Retrieval (TBIR)
\cite{chen2010tag, liu2011textual}. TBIR is effective in achieving
semantic similarity \cite{datta2008image} due to the rich semantic
information possessed by the text data. However, with the exponential
growth of web images, one cannot assume all images on the Internet
have the associated textual data. Clearly, TBIR is not
able to retrieve untagged images even if they are semantically relevant
to the input text query. 

On the other hand, Content Based Image Retrieval (CBIR)
\cite{smeulders2000content, datta2005content, lew2006content,
datta2008image} takes an image as the query and searches for relevant
images based on the visual similarities between the query image and the
database images.  Despite a tremendous amount of effort in the last two
decades, the performance of CBIR system is bounded by the semantic gap
between low-level visual features and high-level semantic
concepts. 

To bridge the semantic gap, one idea is to leverage well annotated
Internet images. Due to the rich information over the Internet, numerous
images are well annotated with text information such as tags,
labels, sentences or even paragraphs.  Their visual content provides low-level visual features while their associated textual information provides
meaningful semantic information.  The complementary nature of texts and
images provides more complete descriptions of underlying content. It is
intuitive to combine both image and text modalities to boost the image
retrieval performance, leading to Multimodal Image Retrieval (MIR). 

\begin{figure}[!t]
\centering
\subfloat[A toy MIR database with tagged and untagged images.]
{\includegraphics[width=0.38 \textwidth]{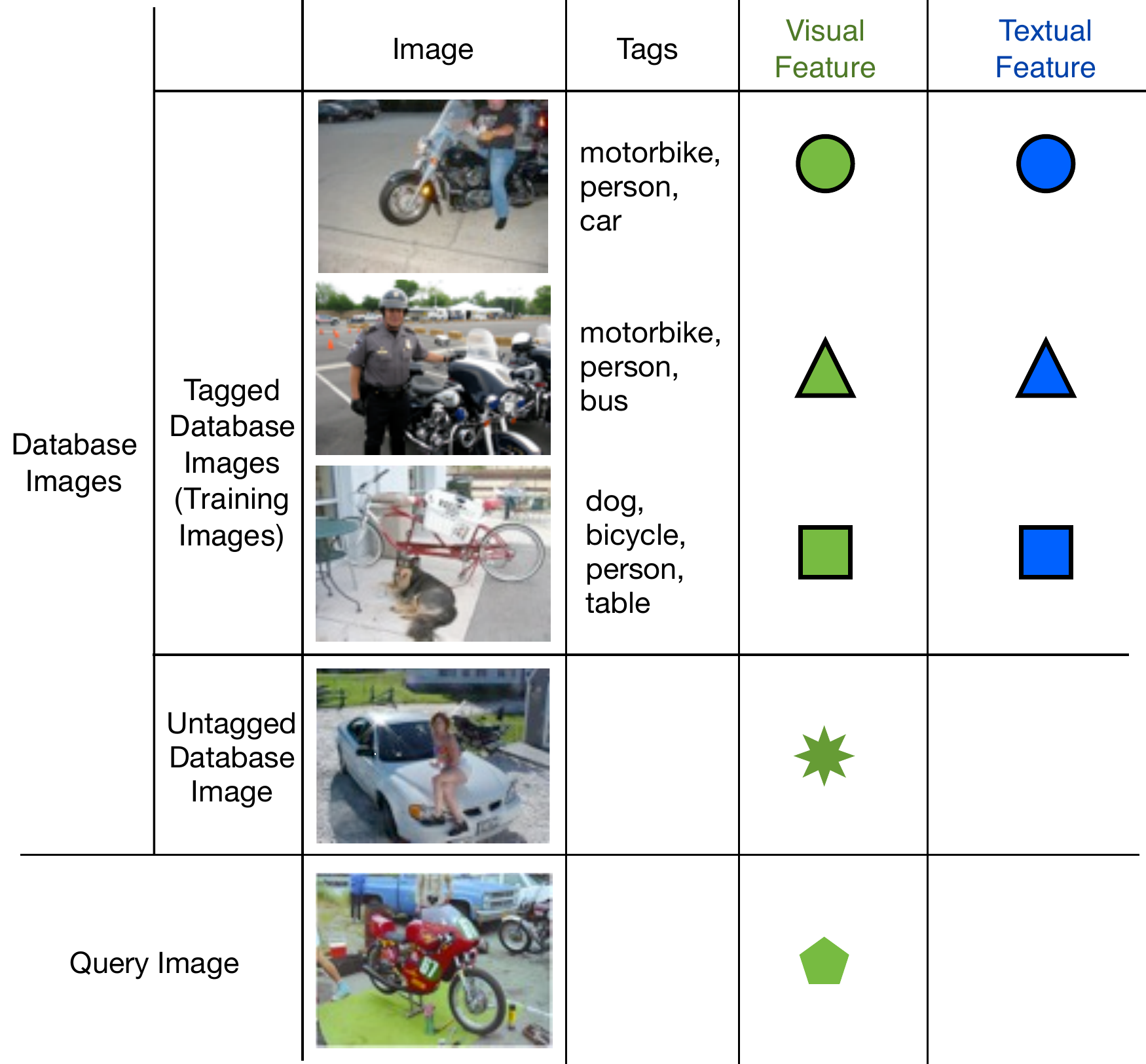}%
\label{fig:MIR_example_table}}
\hfil
\subfloat[Image-to-image search.]
{\includegraphics[width=0.38\textwidth]{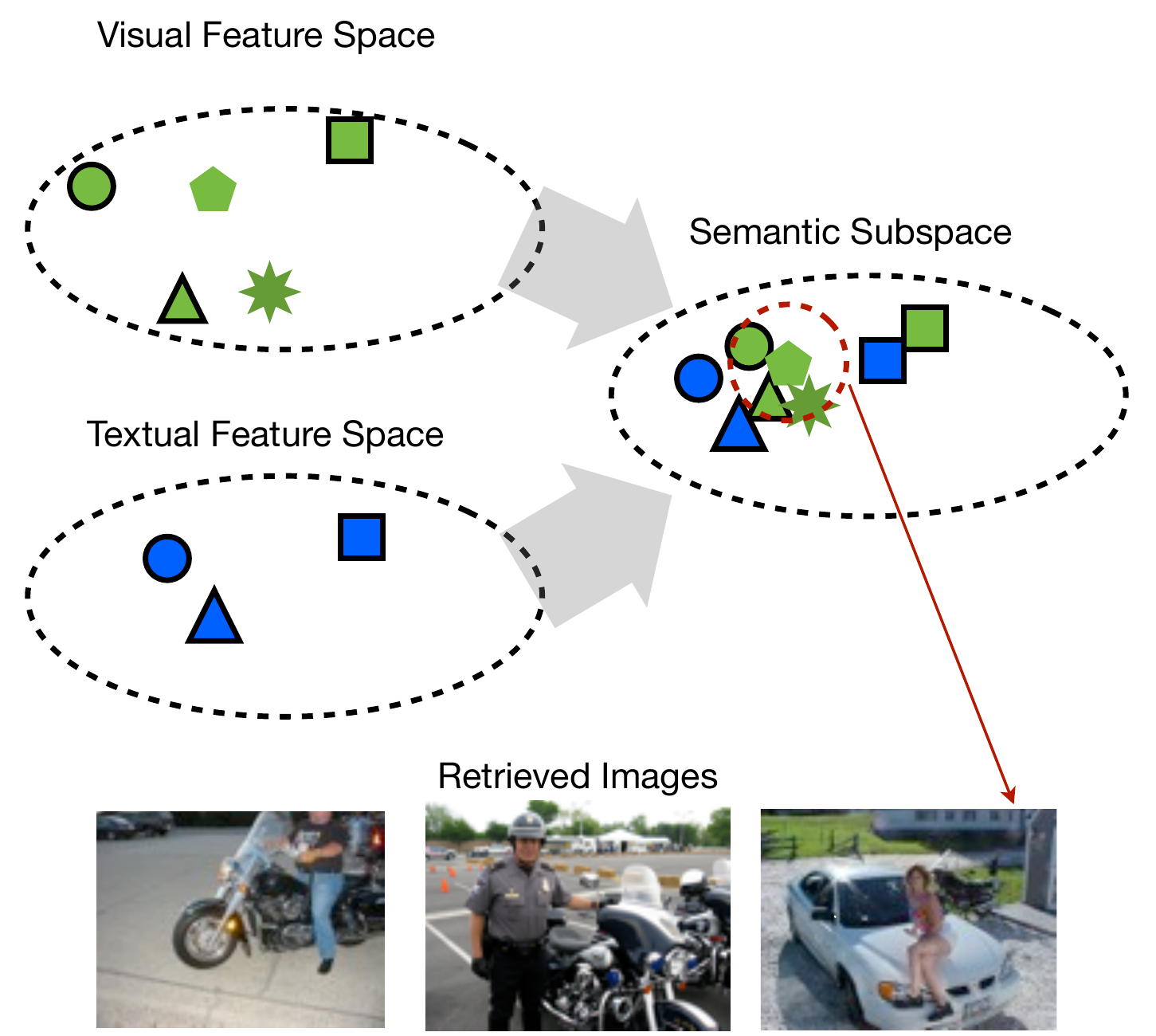}%
\label{fig: MIR_example_framework}}
\caption{Image-to-image search under the MIR framework. (a) A toy
database consisting of both tagged and untagged images. (b) Illustration
of the image-to-image search. Tagged database images are used to learn
the common semantic subspace between the visual and the textual domains
in the training stage.  Visual features of both the query image and
untagged database images will be projected into the common semantic
subspace to calculate the visual similarity during the query time.}
\label{fig:MIR_example}
\end{figure}

When a large number of database images with their textual information are available, one could use MIR to find a
common subspace for the visual and textual features of these tagged
images using machine learning algorithms such as the Canonical
Correlation Analysis (CCA) \cite{hwang2010accounting,
hwang2012learning, gong2014multi, hardoon2004canonical,
rasiwasia2010new, costa2014role} or more recently the Deep Learning
\cite{zhang2014start, andrew2013deep, ngiam2011multimodal,
wang2015effective, johnson2015love} techniques.  Once the common
subspace is found, the visual similarity between the query and
tagged database images in the same subspace can indicate the semantic
similarity, thus reducing the semantic gap to some extent.  Moreover,
untagged database images can be retrieved if their projected visual
features in the subspace are sufficiently close to that of the query image.  An example of MIR applied to
image-to-image retrieval is shown in \figurename~\ref{fig:MIR_example}.
Another advantage of the MIR framework is that it can accommodate
different retrieval tasks at the same time. Since the subspace between
the visual and the textual domains has been built, cross-modality
information search, such as text-to-image and image-to-text search
\cite{barnard2003matching, gong2013deep, zhou2015conceptlearner}, can be achieved by leveraging the subspace
\cite{hwang2010accounting, hwang2012learning, gong2014multi}. 

\begin{figure}[t]
\centering
\includegraphics[width=0.7\linewidth]{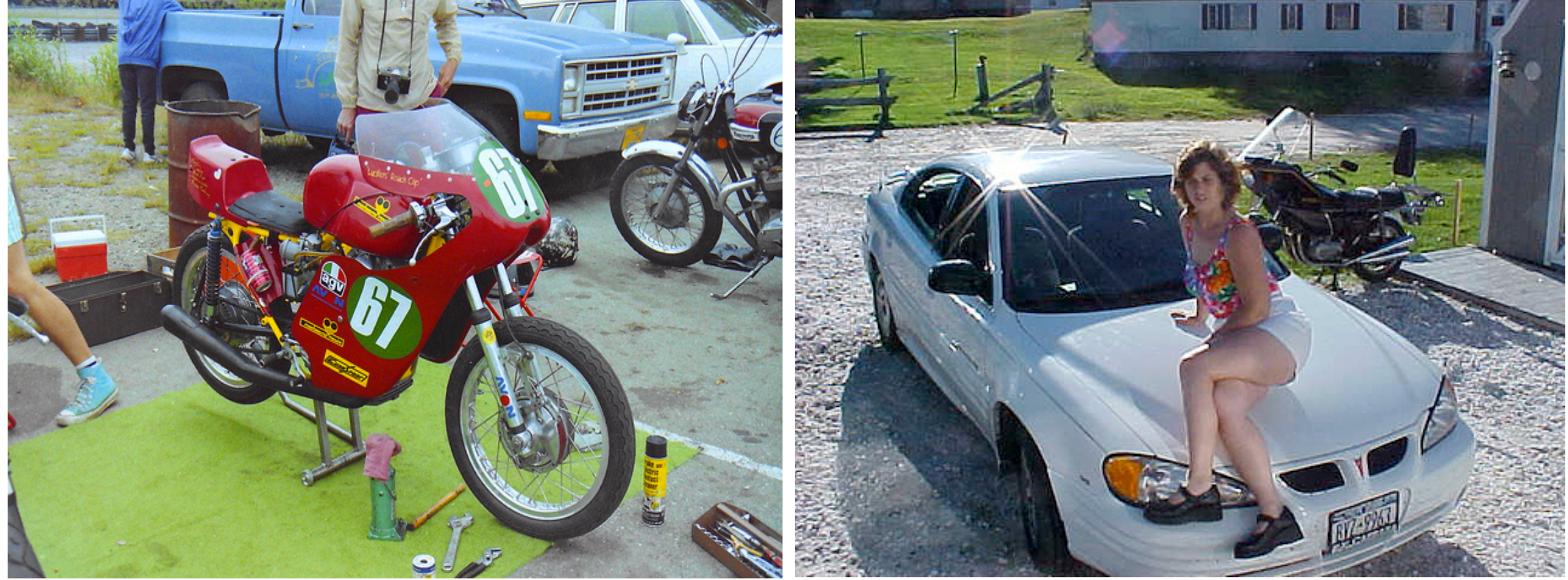}
\caption{Two images with the same object tags (``car", ``motorbike", and
``person") but substantially different visual content.}\label{fig:problem}
\end{figure}

The performance of an MIR system is highly dependent on the quality of tags.
Unfortunately, as pointed out in \cite{wu2013tag, liu2009tag}, tags
provided by people over the Internet are often noisy, and they might not
have strong relevance to image content.  Even if a tag is relevant, the
content it represents might not be perceived as important by humans.
Take \figurename ~\ref{fig:problem} as an example. It has two images
with the same tags: ``car", ``motorbike", and ``person". While the tag
``motorbike" is perceived as having higher importance in the left image,
it is not as important as ``car" and ``person" in the right one.  Thus,
if the left one is the query image, the right one will not be a good
retrieved result. Therefore, capturing and incorporating
human perceived tag importance can significantly improve the performance of
automatic MIR systems, which will be clearly demonstrated in this work. 

When people describe an image, they tend to focus on important content
in the image. Thus, sentence descriptions serve as a natural indicator
of tag importance. Berg {\em et al.} \cite{berg2012understanding}
modeled tag importance as a binary value, i.e.\ a tag is important if
and only if it appears in a sentence. However, when multiple tags are present in the sentences, they might not be equally important. Thus, the binary-valued tag
importance cannot capture the relative tag importance, which degrades
the retrieval performance of MIR systems. To address this deficiency, we
study the relative tag importance prediction problem and incorporate the predicted
importance in an MIR system, leading to the MIR with Tag Importance
Prediction (\textbf{MIR/TIP}) system. To the best of our knowledge, this is the
first work that embeds predicted tag importance into the semantic subspace learning
in an MIR system. 

It is worthwhile to mention that tags can be rather versatile in the wild. On one hand, it may refer to object (``person") or scene (``beach"), which is closely related to the image visual content. On the other hand, it may also refer to more abstractive content such as attribute(``joy") or action (``surfing"). Last but not least, it may refer to photography techniques or camera parameters. The different properties of these content make it hard to study all types of tags at the same time. Thus, in this paper, we focus on studying the importance of object and scene tags. Specifically, we assume all tags are relevant to image visual content,
and do not consider irrelevant (or noisy) tags since
tag cleaning is a research topic on its own. Based on this assumption, we treat image object/scene category labels as the tags in our experiments.

To build an MIR/TIP system, we need to address the following questions:
\begin{enumerate}
\item How to define tag importance?
\item How to predict the defined tag importance?
\item How to embed tag importance in MIR?
\end{enumerate}
For the first question, we propose a method to measure the object and
scene tag importance from human provided sentence descriptions based on
natural language processing (NLP) tools.  A subjective test is conducted
to validate its benefits in image retrieval. For the second question, we
present a novel prediction model that integrates visual, semantic, and
context cues. While the first two cues were explored before in
\cite{berg2012understanding, spain2011measuring}, context cue has
not been considered for tag importance prediction.  The context
cue contributes to tag importance prediction significantly, and results
in improved image retrieval performance as demonstrated in our experiments. To train the prediction model,
we use the Structural Support Vector Machine (SSVM) \cite{tsochantaridis2004support} formulation. For the
third question, Canonical Correlation Analysis (CCA) \cite{hardoon2004canonical} is
adopted to incorporate the predicted tag importance in the proposed MIR/TIP
system. 

The rest of this paper is organized as follows. Related previous work
and an overview of the proposed MIR/TIP system are presented in
Section~\ref{related} and Section~\ref{Overview}, respectively.  A
technique to measure tag importance based on human sentence descriptions
is discussed in Section~\ref{measurement}.  Tag importance prediction is
studied in Section~\ref{predict}, where the measured tag importance is
used as the ground truth.  The MIR/TIP system is described in
Section~\ref{KCCA}. Experimental results are shown in
Section~\ref{results}. Finally, concluding remarks are given and future
research directions are pointed out in Section~\ref{conclusion}. 

\section{Related Work} \label{related}

In this section, we review recent work on tag importance prediction and
MIR systems that are closely related to our work. 

\textbf{Importance Prediction.} Importance in images is a
concept that has recently gained attention in visual research community. Elazary
and Itti \cite{elazary2008interesting} used the naming order of objects
as the interestingness indicator and saliency to predict their locations
in an image. A formal study of object importance was conducted by Spain
and Perona \cite{spain2011measuring}, who developed a forgetful urn model to
measure object importance from ordered tag lists, and then,
used visual cues to predict the object importance value.  Berg {\em
et al.} \cite{berg2012understanding} used human sentence descriptions to
measure the importance of objects, scenes and attributes in images, and
proposed various visual and semantic cues for importance prediction.
To better understand user defined content importance, Yun {\em et
al.} \cite{yun2013studying} studied the relationship between the human
gaze and descriptions. Parikh {\em et al.} \cite{parikh2011relative,
turakhia2013attribute} proposed a ranking method for the same attribute
across different images to capture its relative importance in multiple
images.  Instead of predicting tag importance for images directly, some
works focus on image tags reranking to achieve better retrieval
performance. For example, Liu {\em et al.} \cite{liu2009tag} developed a
random walk-based approach to rerank tags according to their relevance
to image content. Similarly, Tang {\em et al.} \cite{li2012tag} proposed
a two-stage graph-based relevance propagation approach. Zhuang and Hoi
\cite{zhuang2011two} proposed a two-view tag weighting approach to
exploit correlation between tags and visual
features.  Lan and Mori \cite{lan2013max} proposed a Max-Margin Riffled
Independence Model to rerank object and attribute tags in an image in
order of decreasing relevance or importance.  More recently, to address
a very large tag space, Feng {\em et al.} \cite{feng2015learning} cast
tag ranking as a matrix recovery problem. However, all the previous approaches used ranked tag list, human annotators, or binary labels for defining tag importance and failed to capture the continuous-valued relative tag importance which is addressed in our paper.

\textbf{Multimodal Image Retrieval (MIR).} The current state-of-the-art
MIR systems aim at finding a shared latent subspace between image visual
and textual features so that the information in different domains can be
represented in a unified subspace. Several learning methods have been
developed for this purpose, including the canonical correlation analysis
(CCA) \cite{hotelling1936relations} and its extension known as the kernel CCA (KCCA) \cite{lai2000kernel}.  The main idea
of CCA is to find a common subspace for visual and textual features so
that their projections into this lower dimensional representation are
maximally correlated.  Hardoon {\em et al.} \cite{hardoon2004canonical}
adopted KCCA to retrieve images based on their content using the text
query.  Rasiwasia {\em et al.} \cite{rasiwasia2010new} replaced the
textual modality with an abstract semantic feature space in KCCA
training.  More recently, Gong {\em et al.} \cite{gong2014multi}
proposed a three-view CCA that jointly learns the subspace of visual, tag and
semantic features. Hwang and Grauman \cite{hwang2010accounting,
hwang2012learning} adopted human provided ranked tag lists as object tag
importance and used them in KCCA learning, which is most relevant to our
work. Deep learning based multimodal methods adopt the same idea of
learning the shared representation of multi-modal data. However, they
are based on recently developed deep learning techniques such as
stacked autoencoder \cite{ngiam2011multimodal, zhang2014start,
wang2015effective} and deep convolutional neural network
\cite{wang2015effective, johnson2015love}.  The quality of the learned
semantic subspace with shared representations of visual and textual
modalities highly depends on the quality of tags. A noisy or unimportant
tag may lead to misalignment between the visual and the textual domains,
resulting in degenerated retrieval performance. Thus, there is a need to systematically
build an MIR system by incorporating the learned tag importance model, which is one of the focuses of this paper. 

\section{System Overview} \label{Overview}

\begin{figure*}[t]
\centering
\includegraphics[width=1.0\linewidth]{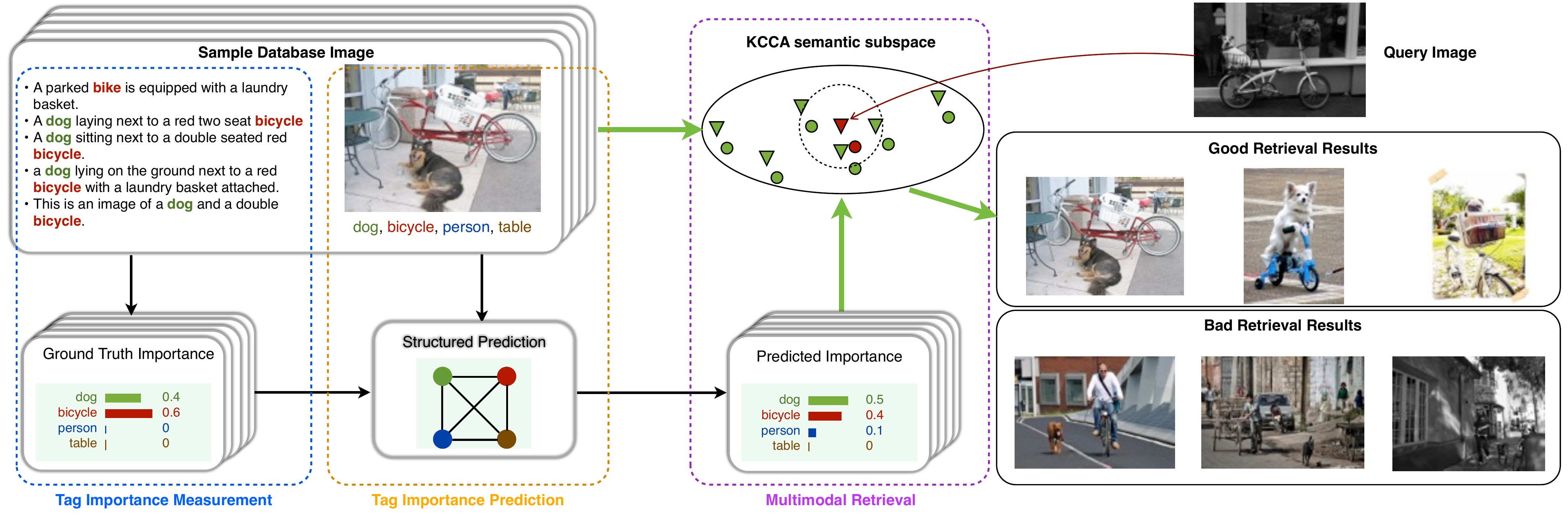}
\caption{An overview of the proposed MIR/TIP system. Given a query image
with important ``dog" and ``bicycle", the MIR/TIP system will rank the
good retrieval examples with important ``dog" and ``bicycle" ahead of
bad retrieval ones with less important ``dog" and ``bicycle".}
\label{fig:framework}
\end{figure*}

A high-level description of the proposed MIR/TIP system is given in
\figurename~\ref{fig:framework}. It consists of the following three
stages (or modules):
\begin{enumerate}
\item Tag importance measurement;
\item Tag importance prediction; and
\item Multimodal retrieval assisted by predicted tag importance. 
\end{enumerate}
It is assumed in our experiments that there are three types of
images on the web:
\begin{enumerate}
\item [A] images with human provided sentences and tags;
\item [B] images with human provided tags only; and
\item [C] images without any textual information.
\end{enumerate}

In the tag importance measurement stage, images in Type A are used to
obtain the measured tag importance, which will serve as the ground truth
tag importance. In the tag importance prediction stage, images in Type A
are first used as the training images to create a tag importance
prediction model. Then, tag importance for images in Type B will be
predicted based on the learned model.  Finally, images in Types A and B
will be used as training images to learn the CCA semantic subspace in
the multimodal retrieval stage. Images in Type C will serve as test
images to validate the performance of our MIR/TIP system.  
Table~\ref{feature_summary} summarizes the textual and visual features used in different stages of MIR/TIP.
Details of each module
will be described in the following sections. 
\begin{table*}[]
\centering
\caption{Visual and textual features used for tag importance measurement, prediction, and MIR.}
\label{feature_summary}
\begin{tabular}{c|c||c|c}
\hline
\multicolumn{2}{c||}{}                                                                                          & \textbf{Textual}                                                                                            & \textbf{Visual}                                                                                          \\ \hline \hline
\multicolumn{2}{c||}{Tag Importance Measurement (Sec. 4)}                                                                      & Tags, Sentences                                                                                    & NA                                                                                              \\ \hline 
\multirow{2}{*}{Tag Importance Prediction (Sec. 5)} & \begin{tabular}[c]{@{}c@{}}Training Data\\ Image Type A\end{tabular}      & Tags,  Measured Importance                                                                         & \begin{tabular}[c]{@{}c@{}}R-CNN Detected Object Bounding Box, \\ Places VGG16 FC7, Saliency\end{tabular} \\ \cline{2-4} 
                                                   & \begin{tabular}[c]{@{}c@{}}Testing Data\\ Image Type B\end{tabular}       & Tags                                                                                               & \begin{tabular}[c]{@{}c@{}}R-CNN Detected Object Bounding Box, \\ Places VGG16 FC7, Saliency\end{tabular} \\ \hline
\multirow{2}{*}{MIR System (Sec. 6)}               & \begin{tabular}[c]{@{}c@{}}Training Data\\ Image Type A \& B\end{tabular} & \begin{tabular}[c]{@{}c@{}}Tags, (A) Measured Importance, \\ (B) Predicted Importance\end{tabular} & \begin{tabular}[c]{@{}c@{}} ImageNet VGG16 FC7, \\ Places VGG16 FC7\end{tabular}                         \\ \cline{2-4} 
                                                   & \begin{tabular}[c]{@{}c@{}}Testing Data\\ Image Type C\end{tabular}       & NA                                                                                                 & \begin{tabular}[c]{@{}c@{}}ImageNet VGG16 FC7, \\ Places VGG16 FC7\end{tabular}                         \\ \hline
\end{tabular}\end{table*}

\section{Measuring Tag Importance} \label{measurement}

Measuring human-perceived importance of a tag is a critical yet
challenging task in MIR. Researchers attempted to measure the importance
of tags associated with images from two human provided sources: 1)
ranked tag lists \cite{hwang2010accounting, hwang2012learning,
spain2011measuring}, and 2) sentence descriptions
\cite{berg2012understanding}. The major drawback of ranked tag lists is
their unavailability.  Tags are rarely ranked according to their
importance, but rather listed randomly. Obtaining multiple ranked tag
lists from human (using the Amazon Turk) is labor intensive and, thus,
not a feasible solution. In contrast, human sentence descriptions are
easier to obtain due to the rich textual information on the Internet.  For
this reason, we adopt sentence descriptions as the source to measure tag
importance. 

\begin{figure}[!t]
\centering
\includegraphics[width=0.8\linewidth]{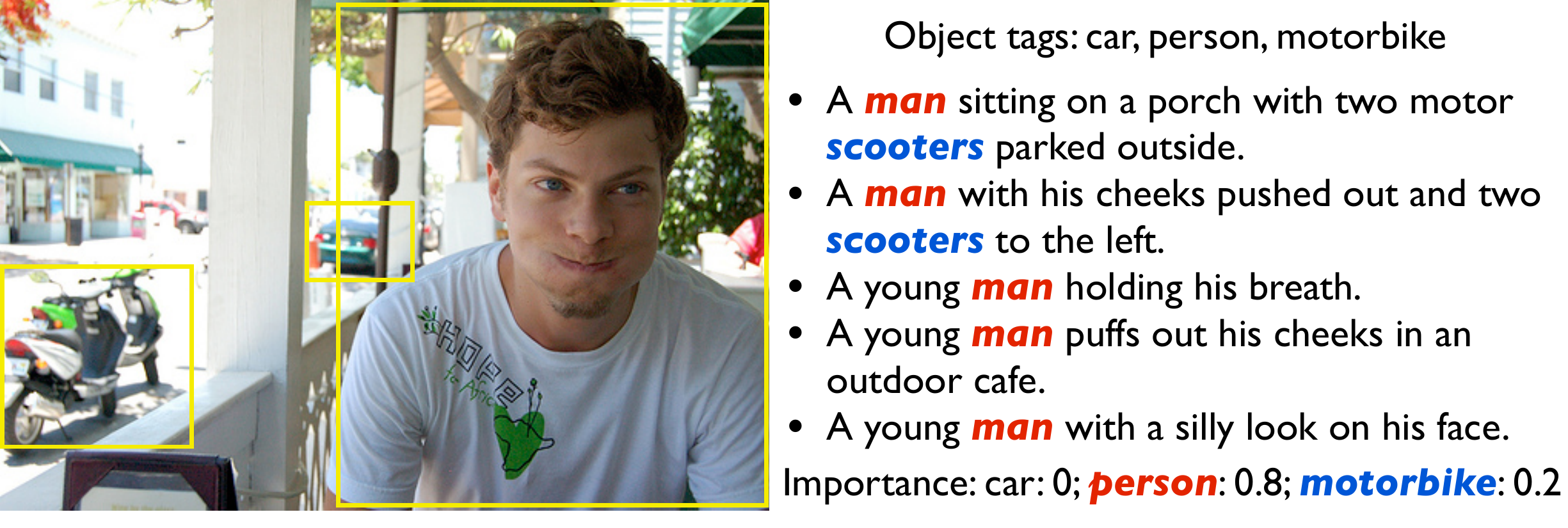}
\caption{An example of object importance measurement using sentence
descriptions, where object tags ``person" and ``motorbike" appear in
sentences in synonyms such as ``man" and ``scooter", respectively.}
\label{fig:measure}
\end{figure}

Clearly, the binary-valued tag importance as proposed in
\cite{berg2012understanding} cannot capture relative tag importance in
an image.  For example, both ``person" and ``motorbike" in
\figurename~\ref{fig:measure} are important since they appear in multiple
sentences. However, as compared with ``person" that appears in all five
sentences, ``motorbike" only appears twice.  This shows that humans
perceive ``person" as more important than ``motorbike" in
\figurename~\ref{fig:measure}.  Thus, tag importance should be
quantified in a finer scale rather than a binary value. 

The desired tag importance should serve the following two purposes.
\begin{enumerate}
\item \textbf{Within-image comparison.} Tag importance should teach the 
retrieval system to ignore unimportant content within an image. 
\item \textbf{Cross-image comparison.} Given two images with the same
tag, tag importance should identify the image in which the tag
has a more important role. 
\end{enumerate}
In the following paragraphs, we will first introduce the idea of measuring object
tag importance and, then, extend it to account for scene tag importance.

\begin{figure}[!t]
\centering
\includegraphics[width=0.7\linewidth]{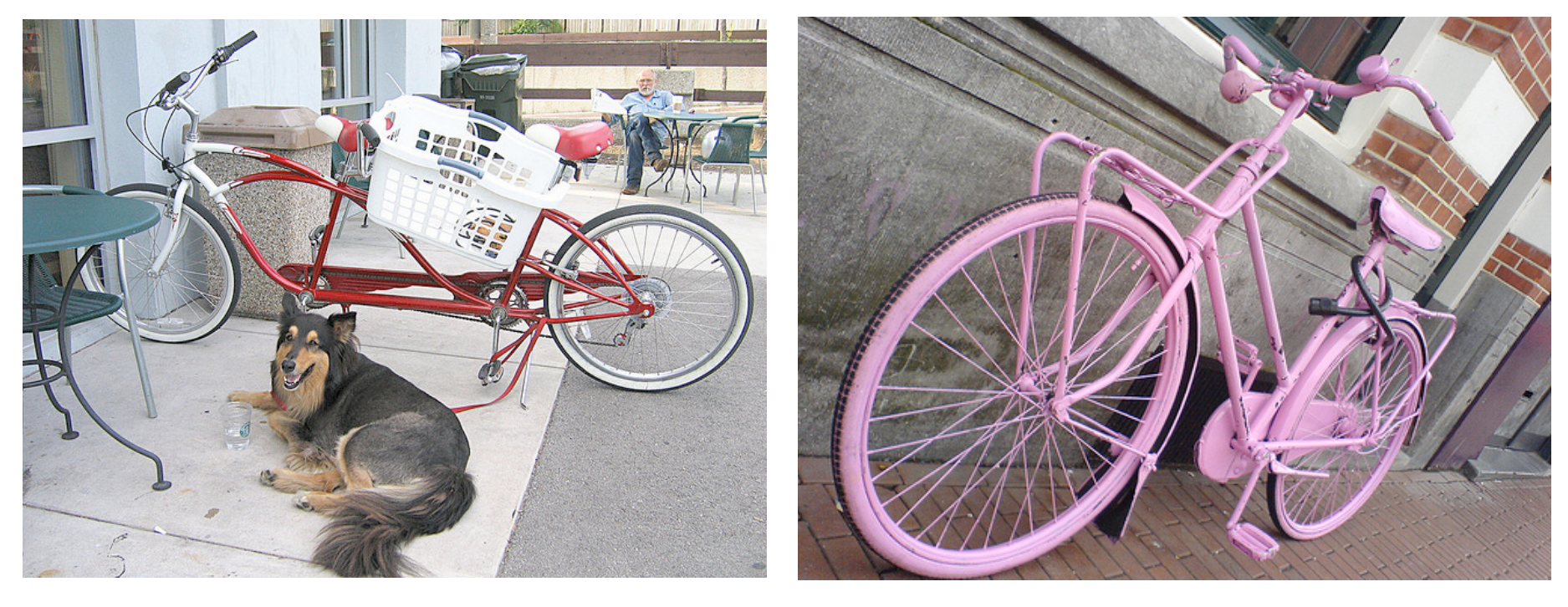}
\vspace{-3mm}
\caption{An example for comparison between probability importance and
discounted probability importance, where the ``bicycle" in both images
are equally important with probability importance but not with the
discounted probability importance (left ``bicycle": 0.6; right
``bicycle": 1.0). }\label{fig:compare}
\end{figure}

\textbf{Object Tag Importance.} To achieve within-image comparison, one
heuristic way is to define the importance of an object tag in an image
as the probability of it being mentioned in a sentence. This is called
\textbf{probability importance}.  To give an example, for the left image
in \figurename ~\ref{fig:compare} (i.e. the sample database image in
\figurename~\ref{fig:framework}), the importance of ``dog" and
``bicycle" are 0.8 and 1 since they appear in four and five sentences,
respectively.  While this notion can handle within-image comparison, it
fails to model cross-image comparison. For instance, as compared with
the right image in \figurename ~\ref{fig:compare}, where the ``bicycle"
is the only tag appearing in all five sentences, it is clear that the
``bicycle" in the left image is less important.  However, its
probability importance has the same value, 1, in both images. 

To better handle cross-image comparison, we propose a measure called
\textbf{discounted probability importance}. It is based on the
observation that different people describe an image in different levels
of detail. Obviously, tags mentioned by detail-oriented people should be
discounted accordingly. Mathematically, discounted probability
importance of object tag $t$ in the $n$th image $\mathbf{I}_n\left ( t
\right )$ is defined as
\begin{equation}\label{eq:importance_eq}
\mathbf{I}_n\left ( t \right )= \sum_{k=1}^{K_n}\frac{\mathbb{I}\left \{
t \in \mathcal{T}_n^{\left ( k \right )} \right \}}{\left |
\mathcal{T}_n^{\left ( k \right )} \right |K_n},
\end{equation}
where $K_n$ is the total number of sentences for the $n$th image,
$\mathbb{I}$ is the indicator function, and $\mathcal{T}_n^{\left ( k
\right )}$ is the set of all object tags in the $k$th sentence of the
$n$th image. An example of measured tag importance using
Eq. (\ref{eq:importance_eq}) is shown in \figurename~\ref{fig:measure}.
Also, for the bicycles in \figurename ~\ref{fig:compare}, the measured
tag importance using discounted probability are 0.6 (left) and 1
(right). 

To identify the appearance of an object tag in a particular sentence, we
need to map the object tag to the word in that sentence. For example,
for the first sentence in Fig.~\ref{fig:measure}, we need to know the
word ``man" corresponds to tag ``person" and ``scooter" corresponds to
tag ``motorbike". We use the WordNet based Semantic distance
\cite{wu1994verbs} to measure the similarity between concepts. The
effectiveness of this synonymous term matching method has been
demonstrated in \cite{berg2012understanding}. 

\textbf{Scene Tag Importance.} To jointly measure scene and object tag
importance, we need to consider two specific properties of the scene tag
in sentence descriptions of an image.
\begin{enumerate}
\item An image usually has fewer scene tags than object
tags. 
\item The grammatical role of the scene tag in a sentence is a strong indicator
of its importance. 
\end{enumerate}

The first property results in an imbalance between the scene and object
tags. Even if both scene and object tags appear in one sentence, scene
importance can be discounted to a lower value if there are many object
tags in the sentence.  To understand the second property, we consider
the following two sentences:
\begin{itemize}
\item Sentence 1: A sandy beach covered in white surfboards near the ocean.
\item Sentence 2: Surfboards sit on the sand of a beach.
\end{itemize}
Clearly, the scene tag ``beach" is the major constituent of the first
sentence because it appears as the main subject of the whole sentence.
On the other hand, it becomes the minor constituent
in the second sentence because it appears in the modifier phrase of
subject ''surfboard". 

\begin{figure}[!t]
\centering
\subfloat[Parse tree of the first sentence]
{\includegraphics[width=0.23 \textwidth]{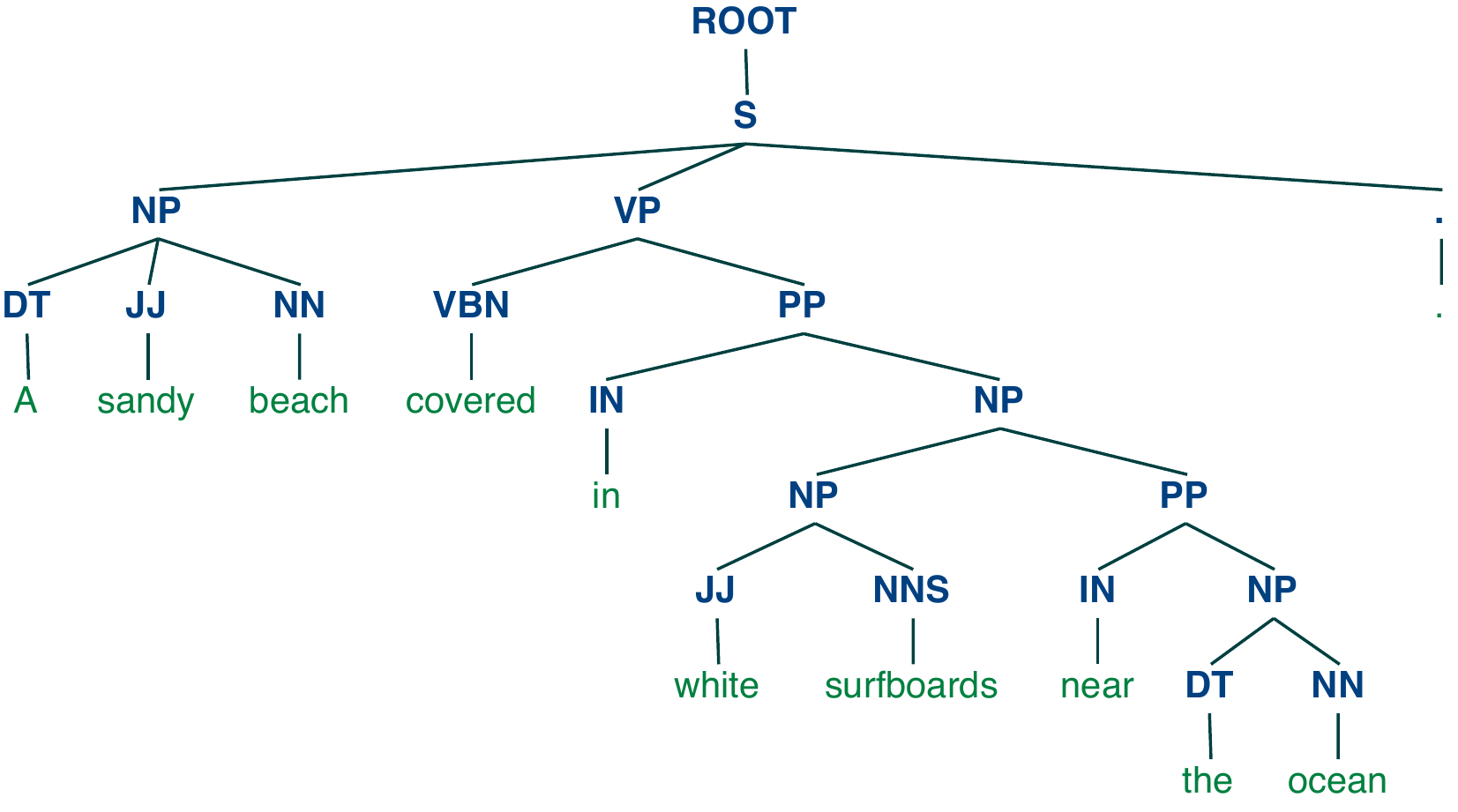}%
\label{fig:tree1}}
\hfil
\subfloat[Parse tree of the second sentence]
{\includegraphics[width=0.23\textwidth]{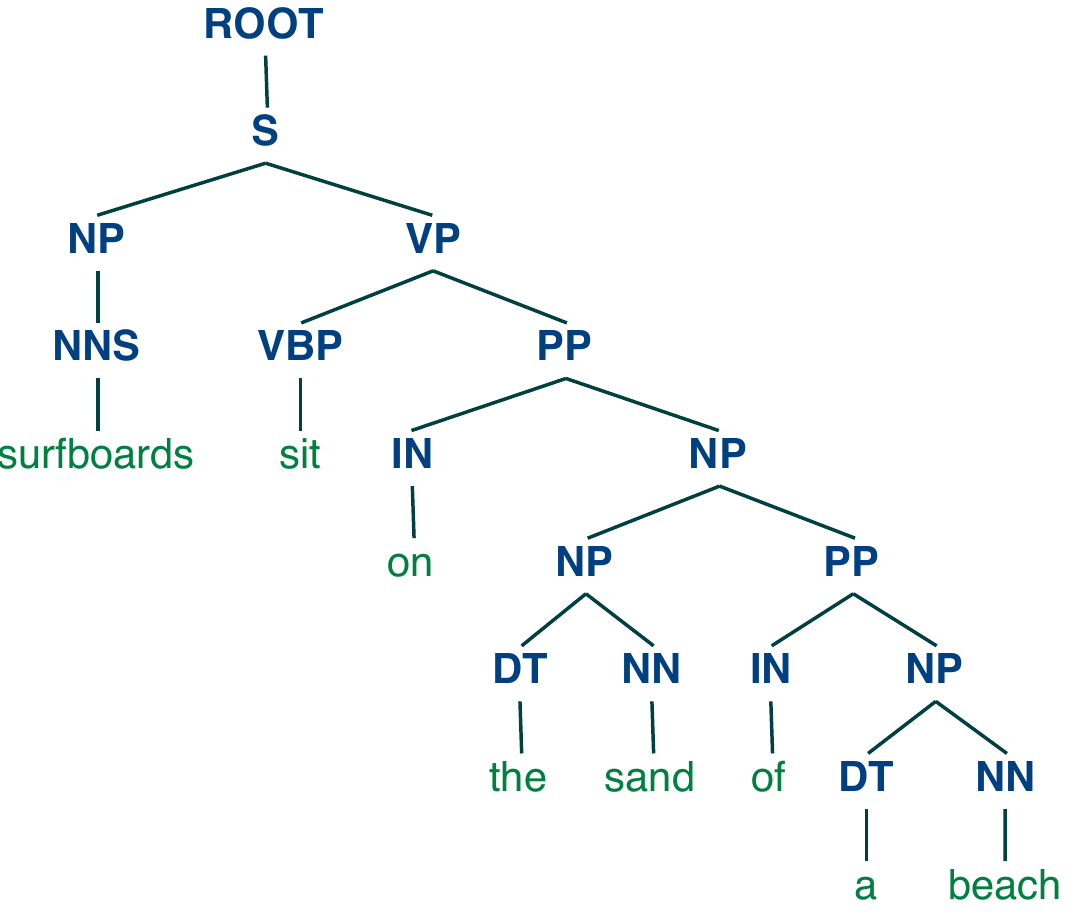}%
\label{fig:tree2}}
\hfil
\caption{The parse trees of two sentences. The
acronyms in the trees are: S (Sentence), NP (Noun Phrase), VP (Verb
Phrase), PP (Preposition Phrase), DT (Determiner), JJ (Adjective), NN
(Singular Noun), NNS (Plural Noun), VBN (Verb, past participle), VBP
(Verb, non-3rd person singular present), IN (Preposition or
subordinating conjunction).} \label{fig:sentence_constituent_structure}
\end{figure}

To infer the grammatical role of a scene tag in a sentence, we first
leverage the Stanford Lexicalized Probabilistic Context Free Grammar
Parser \cite{manning2003natural, manning-EtAl:2014:P14-5} to obtain the
sentence parse tree \cite{BirdKleinLoper09}. Two examples are
shown in \figurename~\ref{fig:sentence_constituent_structure}. Then, we
identify whether the scene tag appears in a prepositional phrase by
checking whether there is a ``PP" node in the path from the root to the
scene tag leaf. 

\textbf{Joint Object/Scene Tag Importance.} Based on the above analysis,
we propose an algorithm to jointly measure the importance of object and scene
tags in one sentence. It is summarized in Algorithm I. The final
scene and object tag importance values are obtained by averaging the
importance vector $\mathbf{I}$ over all sentences associated with the
same image. 

Parameters $\alpha$ and $\beta$ in Algorithm I are the scene weights when
the scene tag appears as the modifier and the subject of a
sentence, respectively. They are used to account for the different
grammatical roles of the scene tag in a sentence. We set $\alpha=1$ and
$\beta=2$ in our experiments.  The object and
scene tag importance values are computed in lines 7 and 8 of Algorithm
I, respectively. The formula in line 7 is essentially the discounted
probability defined in Eq.  (\ref{eq:importance_eq}). The scene tag is adjusted to
account for imbalance of scene and object tag numbers.  

\noindent \textbf{Algorithm I: Measuring object/scene tag importance 
in a sentence description.}
\begin{algorithmic} [1]                   
    \STATE Input: The set of all object tags $\mathcal{T}_o$, the set of all object tags 
     mentioned in sentence $\mathcal{T}_o^{s}$, current sentence $s$, scene tag $t_s$.
    \STATE \textbf{Set} scene factor $\; c_s$ to default value $0$;
    \STATE \textbf{Set} sentence tree $\; \mathbf{T}$ to $\textup{parseSentence}(s)$;
    \STATE \textbf{Set} path $\; \mathbf{p}$ to $\textup{findPath}(  \mathbf{T},  \mathbf{T}.root, t_s)$.
    \STATE \textbf{If}  $\textup{``PP"}$ in $\mathbf{p}\;$ and $\; \textup{``PP"}.left = prep$: \textbf{Set} $c_s = \alpha$ ;
    \STATE \textbf{Else if} $\mathbf{p} \neq NULL$: \textbf{Set} $c_s = \beta$; 
    \STATE \textbf{For} $t \in \mathcal{T}_o$: $\mathbf{I}\left ( t \right )= \frac{\mathbb{I}\left \{
t \in \mathcal{T}_o^{s} \right \}}{\left | \mathcal{T}_o^{s} \right |\left ( 1+c_s \right )}$;
      \STATE  $\mathbf{I}\left ( t_s \right ) = \frac{c_s}{1+c_s}$.
\end{algorithmic}\label{joint_measure}
In the following, the measured tag importance will serve as the ground
truth for our experiments in tag importance prediction in Section~\ref{results}. 

\section{Predicting Tag Importance}\label{predict}

The problem of tag importance prediction is studied in this section.
First, we discuss three feature types used for prediction. They are
semantic, visual and context cues. Then, we describe a prediction model,
in which inter-dependency between tag importance is characterized by the
Markov Random Field (MRF) \cite{felzenszwalb2010object}.  The model
parameters are learned using the Structural Support Vector Machine
(SSVM) \cite{tsochantaridis2004support}. 

\begin{figure}[!t]
\centering
\subfloat[Object semantic]{\includegraphics[width=0.15 \textwidth]{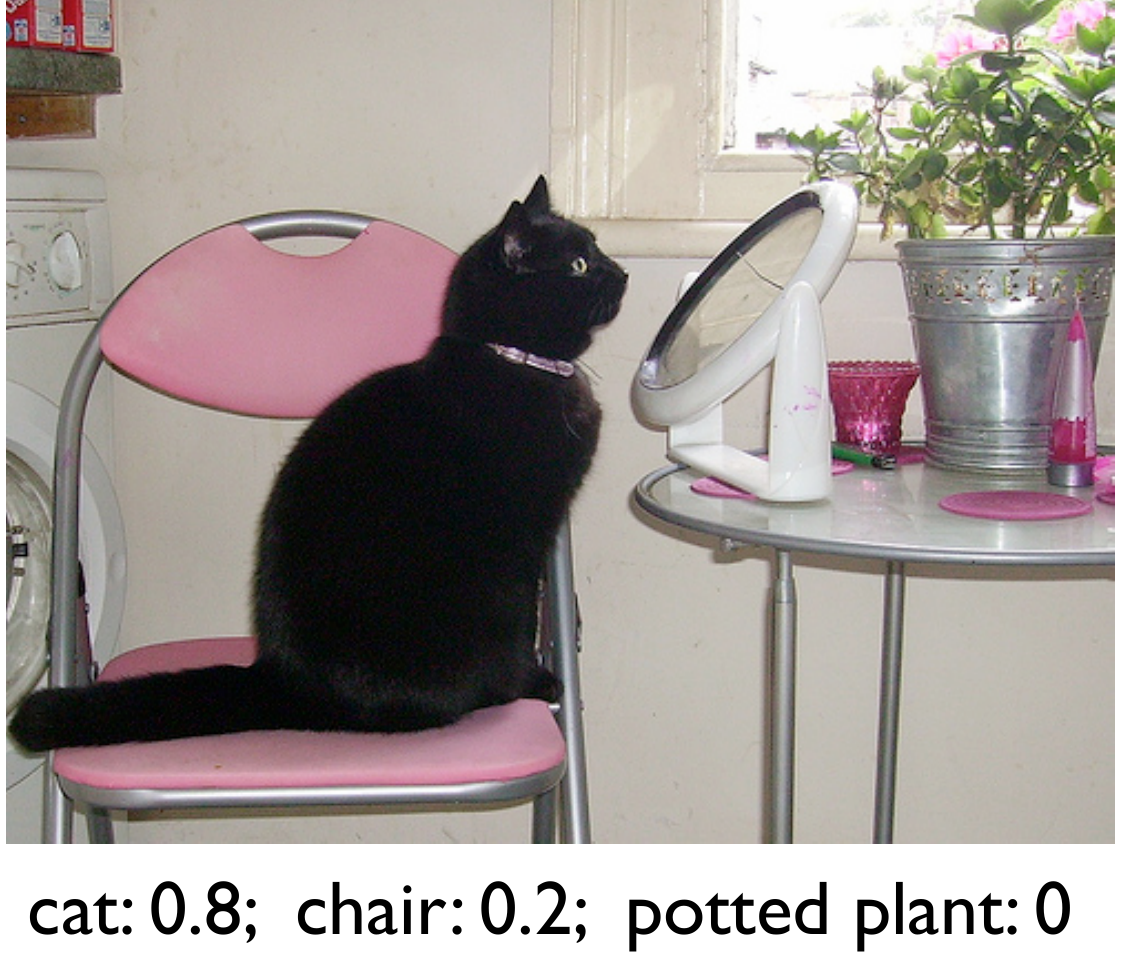}%
\label{fig:semantic}}
\hfil
\subfloat[Object visual]{\includegraphics[width=0.31\textwidth]{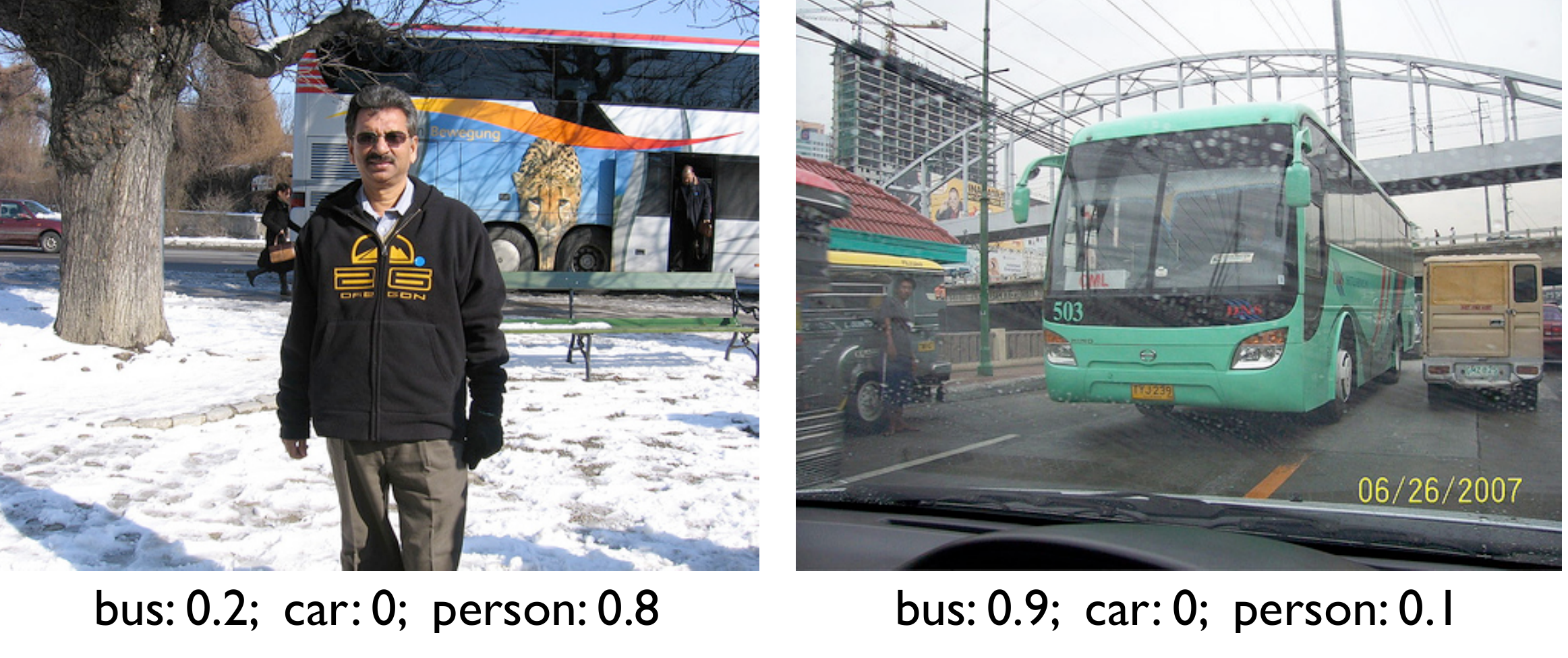}%
\label{fig:visual}}
\hfil
\subfloat[Object context]{\includegraphics[width=0.31\textwidth]{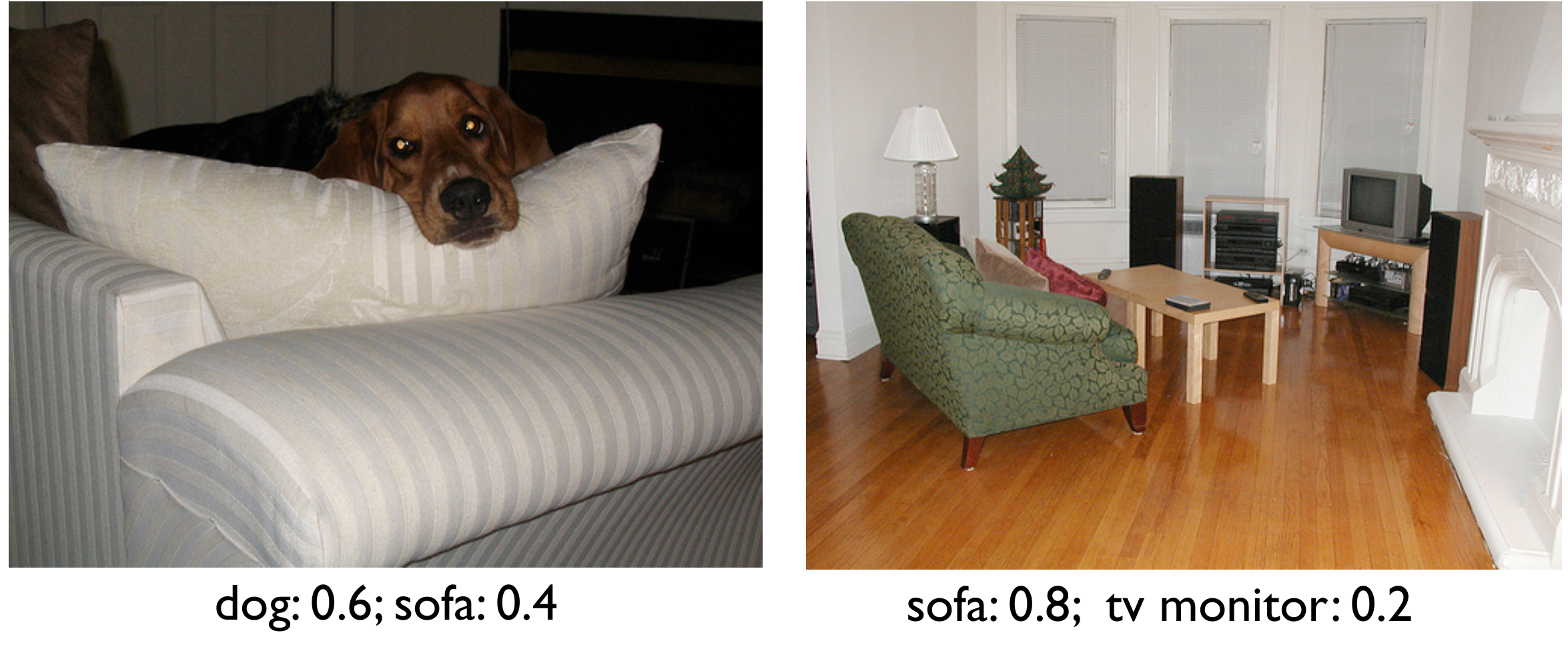}%
\label{fig:context}}
\\ \hfil
\subfloat[Scene semantic]{\includegraphics[width=0.15 \textwidth]{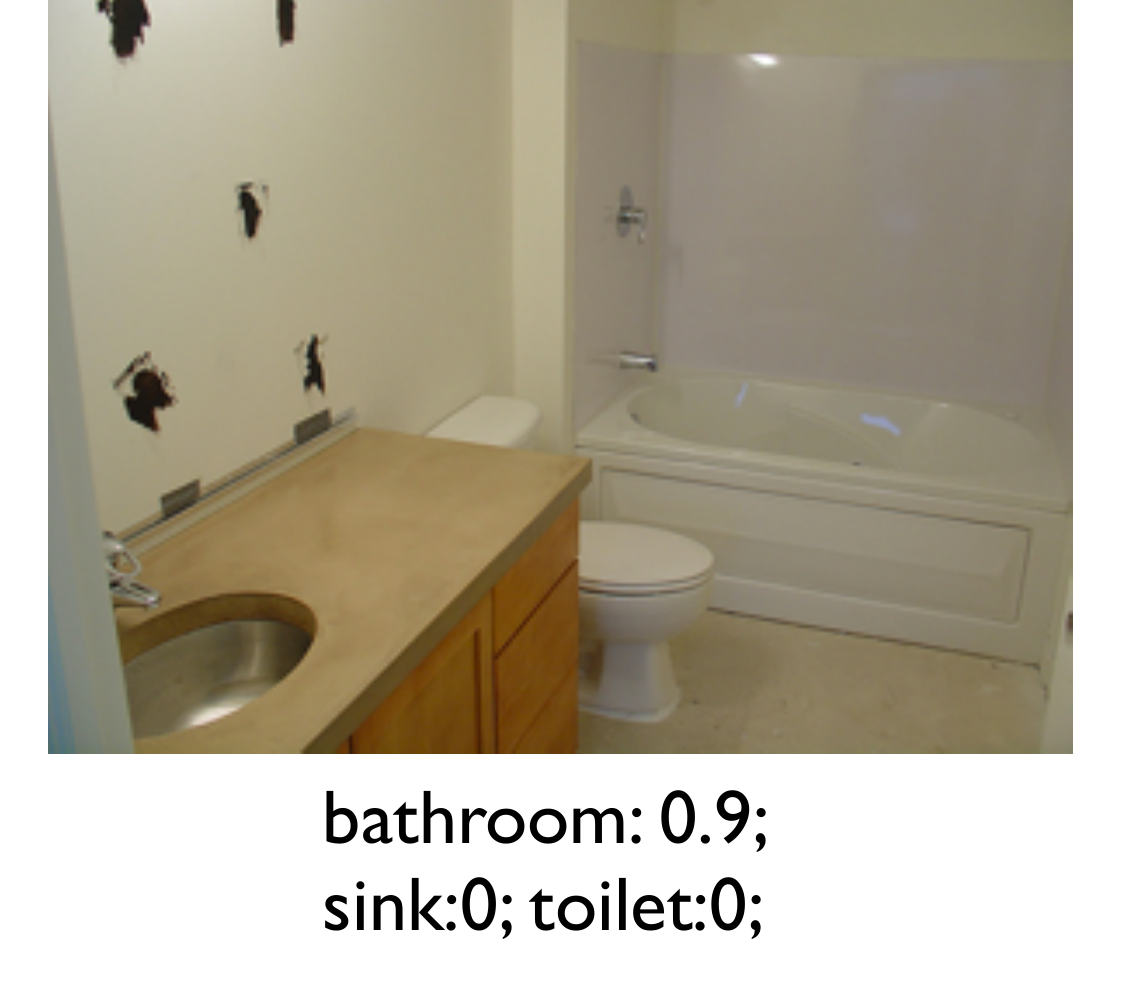}%
\label{fig:scene_semantic}}
\hfil
\subfloat[Scene visual]{\includegraphics[width=0.31\textwidth]{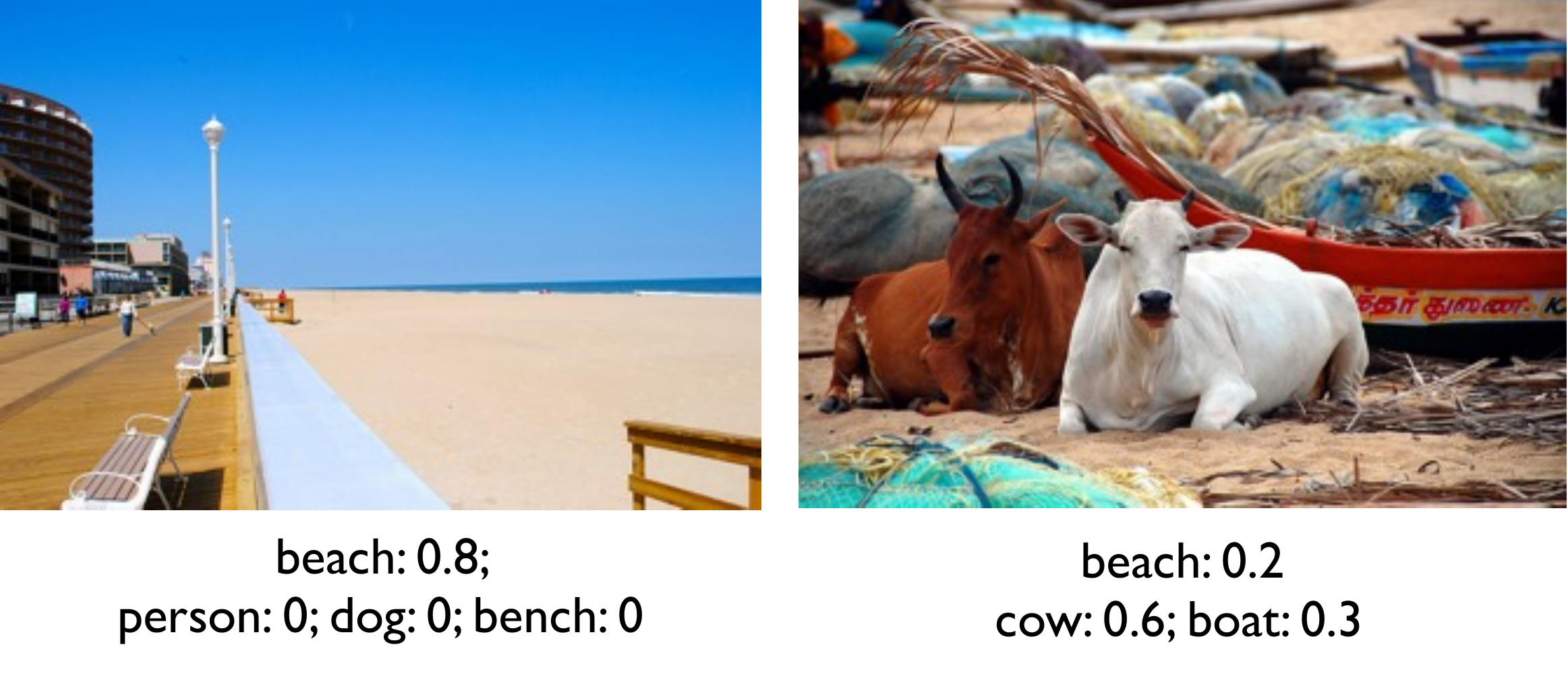}%
\label{fig:scene_visual}}
\hfil
\subfloat[Object scene context]{\includegraphics[width=0.5 \textwidth]{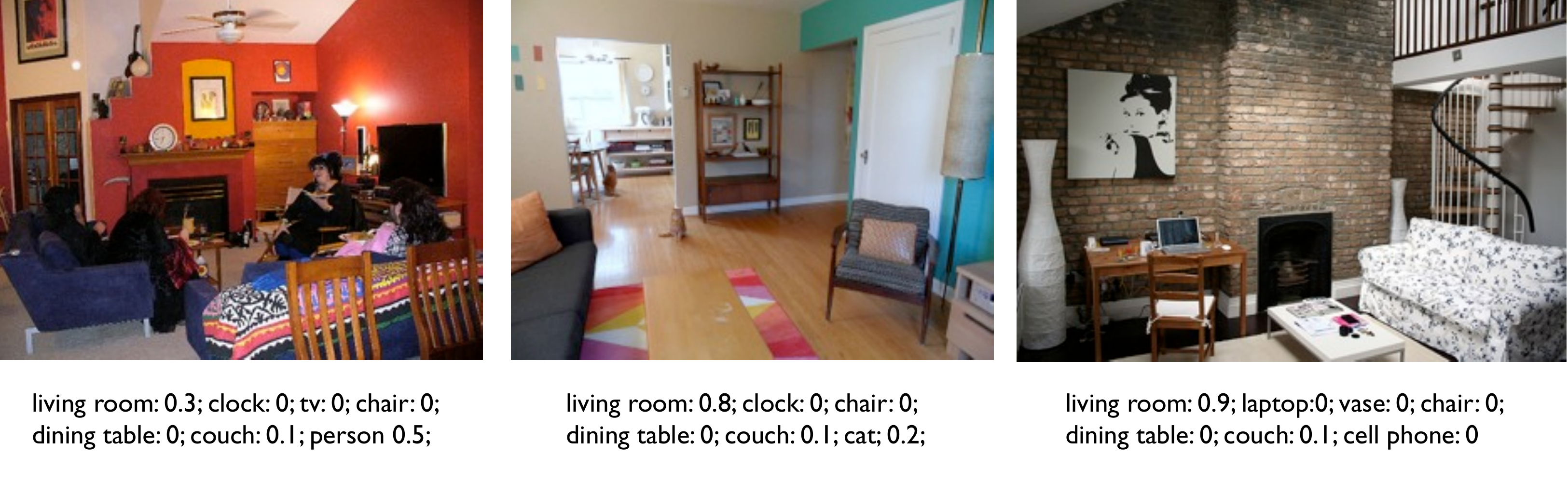}%
\label{fig:scene_context}}
\caption{Examples of various cues for predicting object and scene tag importance.  
The texts below images give the ground truth tag importance.}
\end{figure}

\subsection{Three Feature Types} \label{cue}

\textbf{Semantic Features.} Some object categories are more attractive
to humans than others \cite{berg2012understanding}. For example, given
tags ``cat", ``chair" and ``potted plant" for the image in
\figurename~\ref{fig:semantic}, people tend to describe the ``cat" more
often than the ``chair" and the ``potted plant".  The same observation
applies to scene importance.  
For example, in
\figurename~\ref{fig:scene_semantic}, people often mention (as observed in the dataset) the whole
image as a ``bathroom" image rather than describing objects ``toilet"
and ``sink" in the image. This is because objects ``toilet" and ``sink"
are viewed as necessary components of a bathroom. When all component
objects are combined together to form a scene, people tend to mention
the scene as a whole rather than describe each individual object.  

The semantic cue can be modeled as a categorical feature. It is
a $|\mathcal{C}|$-dimensional binary vector with value one in the $i$th
index indicating the $i$th object/scene category, where $|\mathcal{C}|$ denotes
the number of different object/scene categories.  

\textbf{Visual Features.} Human do not consider an object/scene
important just because of its category. For the case of object tags, we
show a case in \figurename~\ref{fig:visual}, where both images have tags
``bus" and ``person". However, their importance differs because of their
visual properties.  To capture visual cues, we first apply Faster R-CNN \cite{ren2015faster} (with RPN as object proposal network and Fast R-CNN with VGG16 as detector network) to extract object tags' corresponded bounding boxes, and then calculate the following properties using the detected bounding boxes:
1) the area and log(area) as the size features; 2)
the max-, min- and mean-distances to the image
center, the vertical mid-line, the horizontal mid-line, and the third
box \cite{spain2011measuring} as location features; and 3) relative
saliency.  For the last item, we use the spectral residual approach in
\cite{hou2007saliency} to generate the saliency map. Even though false detection will affect the tag importance prediction, it can be reduced significantly by simply removing the proposal whose object category is not in the tag list of current image. The performance loss of tag importance prediction caused by object detection error will be studied in  Section~\ref{results}.
By concatenating all above features, we obtain a 15-D visual feature
vector. An object tag may correspond to multiple object instances in an
image.  For this case, we add the size and saliency features of all
object instances to obtain the corresponding tag size and saliency
features, but take the minimum value among all
related object instances to yield the tag location feature. 

For the scene tag importance visual feature, global patterns such as
``openness" and
color property of an image are useful for predicting scene tag
importance. This is shown in \figurename~\ref{fig:scene_visual}, where
two images have important ``beach" and unimportant ``beach",
respectively. 
We use the FC7 layer (Fully Connected layer 7) features extracted using VGG16 trained on the Places dataset\cite{zhou2014learning} to model the scene property.

\textbf{Object Context Features.} The object context features are used
to characterize how the importance of an object tag is affected by the
importance of other object tags. When two object tags coexist in an
image, their importance is often interdependent.  Consider the sample
database image in Fig.~\ref{fig:framework}.  If the ``dog" did not
appear, the ``bicycle" would be of great importance due to its large
size and centered location, and the discounted probability importance of
``bicycle" would be 1 based on Eq. (\ref{eq:importance_eq}).  To model
interdependency, we should consider not only relative visual properties
between two object tags but also their semantic categories.
Fig.~\ref{fig:context} shows two object context examples.  For the left
image, although the ``sofa" has a larger size and a better location,
people tend to describe the ``dog" more often since the ``dog" gets more
attention. On the other hand, for the right image, people tend to have
no semantical preference between the ``sofa" and the ``TV monitor".
However, due to the larger size of the ``sofa", it is perceived as more
important by humans. 

To extract an object context feature, we conduct two tasks: 1)
analyze the relative visual properties within an object tag pair, and 2)
identify the tag pair type (i.e.\ semantic categories of two object
tags that form a pair).  To model the difference of visual properties
for tag pair $(t_i,t_j)$, we use $s_i-s_j$ and $d_i-d_j$ as the relative
size and location, respectively, where $s$ and $d$ denote
the bounding box area and the corresponding mean distance to the image center
as described in the visual feature section.  The final object context feature $\mathbf{g}^o_{ij}$
for tag pair $(t_i,t_j)$ is defined as
\begin{equation}\label{eq:context_feature_eq}
\mathbf{g}^o_{ij}=\left [ \left ( s_i-s_j \right ) \cdot \mathbf
{p}_{ij}^{\mathsf{T}} \quad \left ( d_i-d_j \right ) \cdot\mathbf
{p}_{ij}^{\mathsf{T}}\right ]^{\mathsf{T}},
\end{equation}
where $\mathbf {p}_{ij}$ is the tag pair type vector for tag pair $(t_i,t_j)$. 

\textbf{Object Scene Context Features.} It is intuitive that scene tag
importance and object tag importance are also interdependent. Consider
three images in \figurename~\ref{fig:scene_context} with the same scene
type - ``living room". The right image is a classical ``living room"
scene and all objects are components of the scene structure. For the
middle image, object ``cat" only takes out a bit of importance of the
``living room" due to its semantic interestingness but relative small
size. At last, in the left most image, people become the dominant
objects due to their prominent size. Thus, the importance of ``living
room" has been suppressed by the tag ``person". Clearly, by removing the
``person" tag in the left image and the ``cat" tag in the middle image,
the ``living room" tags in the three images would be equally important.

To model the context cues between scene and object tag importance, we
use a similar approach described in object context features. That is, we define a
tag pair type vector $\mathbf{p}_{is} $ to indicate the object and scene tags
that form edge $(t_i, t_s)$. It is a categorical vector that models the
semantic of a tag pair. Moreover, as discussed above, the size of an
object may affect the interaction between the scene and object.  Thus,
the context feature is defined as
\begin{equation}\label{eq:scene_context_feature_eq}
\mathbf{g}_{is}^s = s_i\mathbf{p}_{is},
\end{equation}
where $s_i$ is the total size of the $i$th object tag's bounding boxes.

\subsection{Tag Importance Prediction Model}\label{model}
\begin{figure}[!t]
\centering
\includegraphics[width=0.9\linewidth]{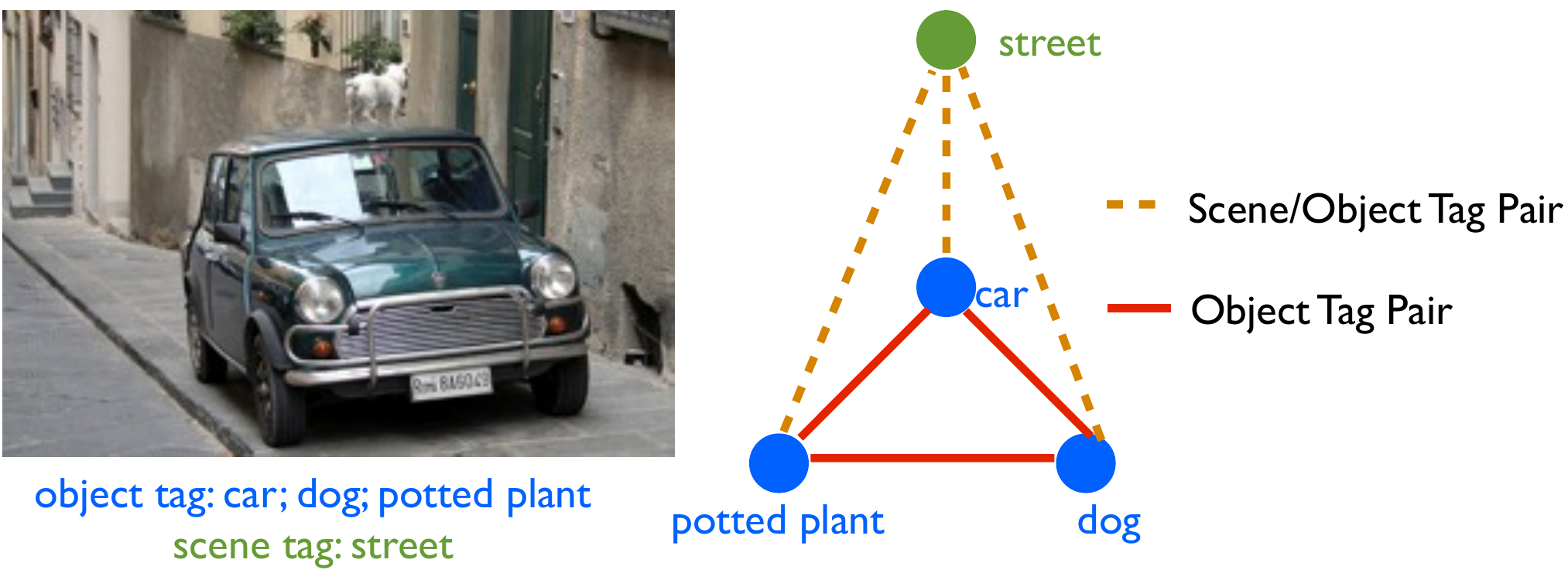}
\caption{A sample image and its corresponding joint MRF model.} 
\label{fig:joint_structured_model}
\end{figure}

The interdependence of tag importance in an image defines a structured
prediction problem \cite{nowozin2011structured}.  It can be
mathematically formulated using the MRF model
\cite{felzenszwalb2010object}.  Formally, each image is represented by
an MRF model denoted by $(V,E)$, where $V=V_o \bigcup v_s $, and where
$V_o$ is the set of object tags of the current image and $v_s$ is the
current scene tag (assuming scene tag is presented in the image). Consider edge set $E=E_o \bigcup E_{os}, $ 
where 
$
E_o = \left \{ \left ( v_i, v_j \right ): v_i, v_j \in V_o \right \}
$ 
is the edge set for object-object tag pairs and 
$
E_{os} = \left \{ (v_i, v_s): v_i \in V_o \right \}
$ 
is the edge set for object-scene tag pairs.  An exemplary MRF model
built for a sample image with scene tag ``street" and object tags
``car", ``dog", ``potted plant" is shown in
\figurename~\ref{fig:joint_structured_model}.  Specifically, each tag
located in a vertex has its own visual and semantic cues to predict
importance while each edge enforces the output to be compatible with the
relative importance between tags. 

Under a log-linear MRF model, the energy function to be minimized
can be expressed as
\begin{align}\label{joint_cost}
\begin{split}
&E(\mathbf{X}_o, \mathbf{x}_s, \mathbf{G}_o,  \mathbf{G}_s, \mathbf{y}; \mathbf{w})=\\&\sum_{i \in
V_o}\underbrace{\mathbf{w}_{V_o}^{\mathsf{T}} \boldsymbol{\varphi}_{V}(\mathbf{x}_i, y_i)}_{\textup{object tag visual \& semantic } } 
+\sum_{\left ( i,j
\right )\in E_o}\underbrace{\mathbf{w}_{E_o}^{\mathsf{T}}  \boldsymbol{\varphi}_{E}(\mathbf{g}^o_{ij}, y_i, y_j)}_{\textup{object tag pair context}}\\
&+\underbrace{\mathbf{w}_{V_s}^{\mathsf{T}} \boldsymbol{\varphi}_{V}(\mathbf{x}_s, y_s)}_{\textup{scene tag visual \& semantic}}
+\sum_{\left ( i,s
\right )\in E_{os}}\underbrace{\mathbf{w}_{E_{os}}^{\mathsf{T}}  \boldsymbol{\varphi}_{E}(\mathbf{g}_{is}^s, y_i, y_s)}_{\textup{object scene tag pair context}},
\end{split}
\end{align}
where $\mathbf{y}=\left \{y_i \right \}$ is the predicted tag importance
output vector, $\mathbf{X}_o=\left \{\mathbf{x}_i \right \}$ and $\mathbf{x}_s$ are the
concatenation of object and scene tag visual and semantic feature vectors as described
in Section~\ref{cue}, respectively. $\mathbf{G}_o=\left \{\mathbf{g}^o_{ij} \right \}$
is the object context feature vector calculated using Eq.
(\ref{eq:context_feature_eq}), $\mathbf{G}_s=\left \{\mathbf{g}^s_{is}
\right \}$ is the object scene context feature vector calculated using
Eq. (\ref{eq:scene_context_feature_eq}). 

The weight vector $\mathbf{w} = \left [ \mathbf{w}_{V_o}^\mathsf{T},
\mathbf{w}_{E_o}^\mathsf{T}, \mathbf{w}_{v_s}^\mathsf{T},
\mathbf{w}_{E_{os}}^\mathsf{T} \right ] ^\mathsf{T} $ in Eq.
(\ref{joint_cost}) will be learned from training data.
$\boldsymbol{\varphi}_V$ and $\boldsymbol{\varphi}_E$ are joint kernel
maps \cite{Bakir:2007:PSD:1296180} defined as
$
 \boldsymbol{\varphi}_V(\mathbf{x_i}, y_i) \equiv \mathbf{x_i}\otimes
\boldsymbol{\delta}\left ( y_i \right ), 
 \boldsymbol{\varphi}_E(\mathbf{g}_{ij}, y_i, y_j) \equiv \mathbf{g}_{ij}\otimes
\boldsymbol{\delta}\left ( y_i-y_j \right )
$
where $\otimes$ is the Kronecker product, 
\begin{align*}
\boldsymbol{\delta}\left ( y_i \right ) & \equiv [\mathbb{I}\left \{ y_i=0
\right \}, \mathbb{I}\left \{ y_i=0.1 \right \}, \cdots, \mathbb{I}\left
\{ y_i=1 \right \}], \\
\boldsymbol{\delta}\left ( y_i-y_j \right ) & \equiv [\mathbb{I}\left \{
y_i-y_j=-1 \right \}, \cdots, \mathbb{I}\left \{ y_i-y_j=1 \right \}],
\end{align*}
where $\mathbb{I}$ is the indicator function. It is worthwhile to mention
that the ground truth tag importance usually takes certain discrete
values in experiments, and it does not affect the retrieval performance
if it is rounded to the nearest tenth. Thus, the ground truth tag
importance is quantized into 11 discrete levels, starting from 0 to 1
with intervals of 0.1. This leads to an 11-D $\boldsymbol{\delta}\left (
y_i \right )$ vector and a 21-D $\boldsymbol{\delta}\left ( y_i-y_j
\right )$ vector.  There are two reasons to define $\boldsymbol{\delta}\left (
y_i-y_j \right )$ such a form.  First, relative
importance can be quantified. Second, dimensionality of the vector $\mathbf{w}$
can be greatly reduced. 

The above model can be simplified to yield the binary-valued tag importance
as done in \cite{berg2012understanding}.  This is achieved by treating
$y_i$ as a binary class label and redefining $\boldsymbol{\delta}\left (
y_i \right )$ and $\boldsymbol{\delta}\left ( y_i-y_j \right )$. The
performance of this simplified model will be reported in the first half
of Section \ref{subsection:TIP}. 

\textbf{Learning.} To learn model parameter $\mathbf{w}$, one
straightforward way is to apply the probabilistic parameter learning
approach \cite{nowozin2011structured}, i.e., treating the energy
function in Eq. (\ref{joint_cost}) as the negative of the log likelihood
of data and applying gradient-based optimization. However, this learning
approach ignores the ordinal nature of the output importance label. For
example, if a tag has ground truth importance value 1, the predicted
importance 0 will be penalized the same as the predicted importance 0.9.
This clearly deviates from intuition. On the other hand, the Loss Minimizing
Parameter Learning approach such as the Structural Support Vector
Machine (SSVM) \cite{tsochantaridis2004support} allows a customizable
loss function for different prediction tasks. It can be exploited by
taking the ordinal nature of output importance label into account.  As a
result, we use the SSVM to learn weight vector $\mathbf{w}$ and adopt
one slack variable with the margin rescaling formulation in
\cite{joachims2009cutting}. The optimization problem becomes
\begin{align}
\begin{split}
\underset{\mathbf{w},\xi}{\min} & \frac{1}{2}\|\mathbf{w}\|^2 + C \xi \\
\textrm{s.t.} & \forall \left ( \bar{\mathbf{y}}_1, \cdots,
\bar{\mathbf{y}}_N \right ) \in \mathcal{Y}^N\\
&\frac{1}{N}\sum_{n=1}^{N}\mathbf{w}^\mathsf{T}
\boldsymbol{\delta}_n\left (\bar{\mathbf{y}}_n \right ) \geq 
\frac{1}{N}\sum_{n=1}^{N}\Delta(\hat{\mathbf{y}}_n, 
\bar{\mathbf{y}}_n)-\xi
\end{split}
\end{align}
where $\hat{\mathbf{y}}_n$ is the ground truth tag importance vector
for the $n$th training image, $\mathcal{Y}^N$ is the set of all possible
output for the training dataset, $C$ is the regularization parameter and $\xi$ is the slack variable, $\boldsymbol{\delta}_n \left ({\bar{\mathbf{y}}}_n \right )$ is the $\boldsymbol{\delta} \left ({\bar{\mathbf{y}}} \right )$ for $n$th image where
\begin{equation*}
\boldsymbol{\delta} \left ({\bar{\mathbf{y}}} \right )
=\Psi(\mathbf{X}_{o}, \mathbf{x}_{s},\mathbf{G}_{o}, \mathbf{G}_{s}, \hat{\mathbf{y}}) -
\Psi(\mathbf{X}_{o}, \mathbf{x}_{s},\mathbf{G}_{o}, \mathbf{G}_{s},\bar{\mathbf{y}}),
\end{equation*}
and
$$
\Psi\left (\mathbf{X}_o, \mathbf{x}_s, \mathbf{G}_o,  \mathbf{G}_s, \mathbf{y} \right )=-\begin{bmatrix}
\sum_{i \in V_o} \boldsymbol{\varphi}_V(\mathbf{x}_i, y_i) \\ \sum_{\left ( i,j
\right ) \in E_o} \boldsymbol{\varphi}_{E}(\mathbf{g}^o_{ij}, y_i, y_j) \\
 \boldsymbol{\varphi}_{V}(\mathbf{x}_s, y_s) \\
  \sum_{\left ( i,s
\right ) \in E_{os}} \boldsymbol{\varphi}_{E{}}(\mathbf{g}^s_{is}, y_i, y_s)
\end{bmatrix}.
$$
Then, we get
$$
\mathbf{w}^\mathsf{T}\Psi\left (\mathbf{X}_o, \mathbf{x}_s, \mathbf{G}_o, 
\mathbf{G}_s, \mathbf{y} \right ) = -E(\mathbf{X}_o, \mathbf{x}_s, \mathbf{G}_o, 
\mathbf{G}_s, \mathbf{y}; \mathbf{w}).
$$  
In the weight learning process, we define the following loss function:
\begin{equation}\label{loss}
\Delta(\hat{\mathbf{y}},\bar{\mathbf{y}}) \equiv \frac{1}{\left | V
\right |}\sum_{i \in V}\left | \hat{y}_i -\bar{y}_i \right |,
\end{equation}
which is the Mean Absolute Difference (MAD) between the ground truth and
the predicted tag importance values of one image. Finally, we applied
the standard cutting plane algorithm \cite{joachims2009cutting} to
optimize and obtain the final weight vector $\mathbf{w}$. 

\textbf{Inference.} After learning the weight vector $\mathbf{w}$, we can
determine the vector $\mathbf{y}$ that minimizes Eq.
(\ref{joint_cost}).  Moreover, finding the maximum violated constraint
in the cutting plane training also needs inference. Despite the fully
connected graph structure in the MRF model, the number of tags in an
image is usually limited. As a result, even if we try all possible outputs,
the computational complexity is still acceptable.  Our experimental
results show that inference takes approximately only 0.2s per image in C
on a 2.4GHz CPU 4GB RAM PC. For this reason, we adopt the exact
inference approach in this work, and we will investigate fast inference techniques in future work. 

\section{Multimodal Image Retrieval} \label{KCCA}

In this section, we discuss our MIR/TIP system by employing CCA/KCCA. First, we will review the CCA and KCCA. Then we will describe the visual and textual features used in our MIR/TIP experiments. Note that other learning methods introduced in Section~\ref{related} can also be used in place of CCA/KCCA.

\subsection{CCA and KCCA}

In multimodal image retrieval, an image is associated with both visual
feature vector $\mathbf{f}_v$ and textual feature vector $\mathbf{f}_t$ (e.g.\ the tag vector).
Given these feature pairs $(\mathbf{f}_v^{(i)},\mathbf{f}_t^{(i)})$ for
$N$ images, two design matrices $\mathbf{F}_v \in \mathbb{R}^{N\times D_v}$ and $\mathbf{F}_t \in \mathbb{R}^{N\times D_t}$ can be generated, where the $i$th row in $\mathbf{F}_v$ and $\mathbf{F}_t $ correspond to $\mathbf{f}_v^{(i)}$ and $\mathbf{f}_t^{(i)}$ respectively. 
CCA aims at finding a pair of matrices $\mathbf{P}_v  \in \mathbb{R}^{D_v\times c}$ and $\mathbf{P}_t  \in \mathbb{R}^{D_t\times c}$ that project visual and textual features into a common $c$ dimensional subspace with maximal normalized correlation:
\begin{align}\label{eq:kcca_obj}
\begin{split}
&\underset{\mathbf{P}_v,\mathbf{P}_t}{\textup{max}}\,\textup{trace}\left ( \mathbf{P}_v^\mathsf{T}\mathbf{F}_v^\mathsf{T}  \mathbf{F}_t \mathbf{P}_t\right )\\
\textup{s.t.} \quad& \mathbf{P}_v^\mathsf{T}\mathbf{F}_v^\mathsf{T}\mathbf{F}_v\mathbf{P}_v = \mathbf{I}, \ \mathbf{P}_t^\mathsf{T}\mathbf{F}_t^\mathsf{T}\mathbf{F}_t\mathbf{P}_t = \mathbf{I}.
\end{split}
\end{align}
The above optimization problem can be reduced to a generalized eigenvalue problem \cite{hardoon2004canonical}, and the eigenvectors corresponding to the largest $c$ eigenvalues are stacked horizontally to form $\mathbf{P}_v$ and $\mathbf{P}_v$.

To measure the similarity of projected features in subspace to achieve cross-modality retrieval, we adopted the Normalized CCA metric proposed in \cite{gong2014multi, gong2014improving}. After solving the CCA problem in Eq. (\ref{eq:kcca_obj}), the similarity between visual features $\mathbf{F}_v$ and textual features $\mathbf{F}_t$ will be computed as:
\begin{equation}\label{eq:ncca}
\frac{\left (\mathbf{F}_v\mathbf{P}_v \textup{diag}\left ( \lambda_1^t, \cdots ,\lambda_c^t \right )  \right )\left ( \mathbf{F}_t\mathbf{P}_t \textup{diag}\left ( \lambda_1^t, \cdots ,\lambda_c^t \right )\right )^\textup{T}}{\left \|\mathbf{F}_v\mathbf{P}_v \textup{diag}\left ( \lambda_1^t, \cdots ,\lambda_c^t \right )  \right \|_2 \left \| \mathbf{F}_t\mathbf{P}_t\textup{diag}\left ( \lambda_1^t, \cdots ,\lambda_c^t \right ) \right \|_2},
\end{equation}
where $ \lambda_1, \cdots ,\lambda_c$ correspond to the top $c$ eigenvalues, and $t$ is the power of the eigenvalues (we set $t$=4 as in \cite{gong2014multi, gong2014improving}).

To model nonlinear dependency between visual and textual feature
vectors, a pair of nonlinear transforms, $\Phi_v$ and $\Phi_t$, are used
to map visual and textual features into high dimensional spaces,
respectively. With kernel functions
$$
{\mathbf{K}_m( \mathbf{f}_m^{(i)},\mathbf{f}_m^{(j)})=\Phi_m( \mathbf{f}_m^{(i)} 
)^\mathsf{T}\Phi_m(\mathbf{f}_m^{(j)} )}\quad {m = v,t}
$$ 
$\Phi_v$ and $\Phi_t$ are only computed implicitly. This kernel trick lead to KCCA, which attempts to find the maximally correlated subspace with the two
transformed spaces  \cite{hardoon2004canonical}. However, since the time and space complexity of KCCA is $O(N^2)$, it is not practically applicable to large scale image retrieval. We thus adopt the CCA for the retrieval experiments. 

We also tried out the scalable KCCA proposed in \cite{gong2014multi} by constructing the approximate kernel mapping but find almost no performance improvement in our experiment setting.

\subsection{Retrieval Features}\label{feature_kernel}

Features used in CCA for MIR/TIP and experimental settings are
given below. 

\textbf{Visual Features.} 
To capture both object and scene properties, we use the VGG16 trained on the ImageNet \cite{krizhevsky2012imagenet} and the Places \cite{zhou2014learning} to extract visual features. The output of the FC7 (Fully Connected layer 7) for both networks are concatenated together to form a 8192-D visual feature vector.

\textbf{Textual Features.} We consider 5 types of textual features, i.e.\ 1) the tag vector, 2) the predicted binary-valued tag importance vector, 3) the true binary-valued tag importance vector, 4)
the predicted continuous-valued tag importance vector, and 5) the true continuous-valued tag
importance vector, each of which is used in a retrieval experimental
setting as discussed in Section~\ref{results}. 

\section{Experimental Results}\label{results}

In this section, we first discuss the datasets and our experimental settings. We proceed to present our subjective test results to justify that the ground truth tag importance based on descriptive sentences is consistent with human perception.
Then, we compare the performance of
different tag importance prediction models. Finally, experiments on three retrieval tasks are conducted to demonstrate the superior
performance of the proposed MIR/TIP system.

\subsection{Datasets}\label{dataset}

To test the proposed system, we need datasets that have many annotated
data available, including sentence descriptions, object tags,
object bounding boxes and scene tags. 
Table~\ref{tab:dataset_summary}
lists the profiles of several image datasets in the public domain. Among them, the UIUC, 
COCO, and VisualGenome datasets appear to meet our need the most since they have descriptive
sentences for each image. 
However, the Visual Genome dataset aims to be an ``open vocabulary" dataset with 80,138 object categories but on average only 49 instances for each object category. While the large tag vocabulary size makes it challenging to learn CCA subspace, limited number of instances per category makes Faster R-CNN training difficult. 
We thus adopt the UIUC and COCO datasets to conduct our experiments, and leave the Visual Genome dataset to our future work.
To test our proposed full system with scene tag, we enrich COCO with scene tags and call it \textbf{COCO Scene dataset} (will be discussed later in this section). In the following, we consider 2 types of datasets: 1) the datasets with only object tags (e.g. COCO, UIUC) and 2) the dataset with both object and scene tags (e.g. COCO Scene). 
\begin{table}[htb]
\caption{Comparison of major image datasets.}\label{tab:dataset_summary}
\centering
\begin{tabular}{c||c|c|c|c} \hline
Dataset & Sentences & \begin{tabular}[c]{@{}c@{}} Object \\ Tag\end{tabular} & \begin{tabular}[c]{@{}c@{}}Bounding \\ Box\end{tabular} & \begin{tabular}[c]{@{}c@{}}Scene \\ Tag\end{tabular} \\ \hline \hline
UIUC \cite{rashtchian2010collecting} & Yes & Yes  & Yes &  No \\ \hline
COCO \cite{lin2014microsoft} & Yes & Yes  & Yes &  No \\ \hline
SUN2012  \cite{xiao2010sun}& No & Yes   & Yes &  Yes \\ \hline
LabelMe \cite{russell2008labelme} & No   & Yes &  Yes  & No\\ \hline
ImageNet \cite{deng2009imagenet} & No  & Yes &  Yes  & No\\ \hline
Visual Genome \cite{krishnavisualgenome} & Yes  & Yes &  Yes  & No\\ \hline
\end{tabular}
\end{table}

For the datasets with only object tags, we adopt UIUC and full COCO datasets for small and large scale experiments respectively. Moreover, the UIUC
\cite{rashtchian2010collecting} dataset was used for image
importance prediction in \cite{berg2012understanding} and can serve as a
convenient benchmarking dataset.  Each image in these two datasets is annotated by 5 sentence
descriptions and each object instance in an image is labeled with a
bounding box. The UIUC dataset consists of 1000 images with objects from
20 different categories. The COCO dataset has in total 123,287 images with objects from 80 categories. 
We use object categories as object
tags. Thus, UIUC has approximately 1.8 tags per image while COCO has on average
3.4 tags per image. We mainly use UIUC dataset for importance prediction experiment since it is not a challenging dataset for retrieval.

For dataset with both object and scene tags, we generated an experimental dataset based on images in the COCO dataset. Specifically, we first identified 30 common scene types in the COCO dataset. Then, 50 human workers were invited to manually classify 60,000 images randomly drawn from the COCO dataset into one of 31 groups, which include 30 scene types mentioned above and an extra group indicated as ``Not sure/None of above". This is necessary as the COCO dataset contains a large amount of object centric images, whose scene types are hard to identify even for human. This resulted in a dataset consisting of 25,124 images with a tag vocabulary of 110. It has on average 4.3 tags per image.  We will refer to this dataset as the COCO Scene dataset for the rest of this section. (For the statistics of COCO Scene dataset, please refer to supplementary material.)

\subsection{Retrieval Experiment Settings}\label{settings}

As introduced in Section~\ref{KCCA}, different tag features correspond
to different MIR experiment settings because the semantic subspace
is determined by applying CCA to visual and textual features. We thus compared the following 5 MIR settings:

\begin{itemize}
\item \textbf{Traditional MIR}: Textual features are the binary-valued tag
vectors. This is the benchmark method
used in \cite{hwang2010accounting, hwang2012learning, gong2014multi}. 
\item \textbf{MIR/PBTI}: Textual features are \textbf{P}redicted \textbf{B}inary-valued \textbf{T}ag
\textbf{I}mportance vectors. This corresponds to the predicted importance proposed in \cite{berg2012understanding}. 
\item \textbf{MIR/PCTI}: Textual features are \textbf{P}redicted \textbf{C}ontinuous-valued \textbf{T}ag
\textbf{I}mportance vectors. This is \textit {our proposed system}. 
\item \textbf{MIR/TBTI}: Textual features are \textbf{T}rue \textbf{B}inary-valued \textbf{T}ag
\textbf{I}mportance vectors. This serves as the upper bound for the binary-valued tag importance proposed in \cite{berg2012understanding}.
\item \textbf{MIR/TCTI }: Textual features are \textbf{T}rue \textbf{C}ontinuous-valued \textbf{T}ag
\textbf{I}mportance vectors. This gives the best retrieval performance, which serves 
as the performance bound. 
\end{itemize}
Among the above five systems, the last two are not achievable since they assume the tag importance prediction to be error free.

Moreover, we evaluate our system in terms of 3 retrieval tasks: 
\begin{itemize}
\item \textbf{I2I} (Image to Image retrieval): Given a query image, the MIR systems will project the visual features into the CCA subspace and rank the database images according to Eq. (\ref{eq:ncca}). We also test a baseline retrieval system (\textbf{Visual Only}) that ranks the database images using visual features' Euclidean distance.
\item \textbf{T2I} (Tag to Image retrieval): Given a tag list, the MIR systems will project the tag feature into the CCA subspace and rank the database images according to Eq. (\ref{eq:ncca}). Note our system can support weighted tag list as query as in \cite{gong2014multi}, in which the weights represent the importance of tags. 
\item \textbf{I2T} (Image annotation): Given a query image, the MIR systems will find 50 nearest neighbors in the CCA subspace and use their textual features to generate an average textual feature vectors, based on which the tags in tag vocabulary will be ranked. We also test a baseline tagging system using deep features to find nearest neighbors and their corresponding tag vectors to rank tags. 
\end{itemize}

For all retrieval tasks, we adopt the Normalized Discounted Cumulative Gain (NDCG) as the
performance metric since it is a standard and commonly used metric \cite{hwang2010accounting,
hwang2012learning, lan2013max}. Moreover, it helps quantify how an MIR
system performs. The NDCG value for the top k results is defined as
$
\mathrm{NDCG@k} = \frac{1}{Z}\sum_{i=1}^{k} \frac{2^{r_{i}}-1}{\log_{2}(i+1)},
$
where $r_i$ is a relevance index (or function) between the query and the
$i$th ranked image, and $Z$ is a query-specific normalization term that
ensures the optimal ranking with the NDCG score of 1. The relevance
index measures the similarity between retrieved results and the query in
terms of ground truth continuous-valued tag importance, i.e.\ whether an MIR system can
preserve important content of the query in retrieval results or not.  For I2T retrieval task, the relevance of a tag to the query image is set as its ground truth continuous-valued tag importance. The choice of the relevance index for the other two tasks will be discussed in detail in the next section.

\subsection{Subjective Test Performance of Measured Tag Importance} \label{subsection:subjective_test}

Since ground truth continuous-valued tag importance is used to measure the degree of
object/scene importance in an image as perceived by a human, it is desired to
design a subjective test to evaluate its usefulness.  Here, we would
like to evaluate it by checking how much it
will help boost the retrieval performance.  Specifically, we compare the
performance of two different relevance functions
and see whether the defined ground truth continuous-valued tag importance correlates human
experience better. The two relevance functions are given below. 
\begin{enumerate}
\item The relevance function with measured ground truth tag importance:
\begin{equation} 
r_g(p,q) =  \frac{\left \langle  \textbf{I}_p,  \textbf{I}_q\right 
\rangle}{\left \| \textbf{I}_p \right \|\left\|\textbf{I}_q \right\|},
\end{equation}
where $\textbf{I}_k$ denotes the ground truth continuous-valued importance vector for
image $k$ ($k=p$ or $q$). 
\item The relevance function with binary-valued importance \cite{berg2012understanding} (whether appeared in sentences or not):
\begin{equation} 
r_b(p,q) =  \frac{\left \langle  \textbf{t}_p,  \textbf{t}_q\right 
\rangle}{\left \| \textbf{t}_p \right \|\left\|\textbf{t}_q \right\|},
\end{equation}
where $\textbf{t}_k$ denotes the binary-valued tag importance vector for image $k$ ($k=p$ or $q$). 
\end{enumerate}

In the experiment, we randomly selected 500 image queries from the COCO
Scene dataset and obtained the top two retrieved results with the max
relevance scores using two relevance functions mentioned above.  In the
subjective test, we presented the two retrieved results of the same
query in pairs to the subject, and asked him/her to choose the better one among the two. 
We invited five subjects (one female and four males
with their ages between 25 and 30) to take the test. There were 1500
pairwise comparisons in total. We randomized the order of
two relevance functions in the GUI to minimize the bias.  Moreover, each
subject viewed each query at most once. We made the following
observation from the experiment. As compared with the results using the
relevance function with binary-valued importance $r_b$, the results using our
relevance function  $r_g$ were favored in 1176 times (out of
1500 or 78.4\%). This indicates that the relevance function with
ground truth importance $r_g$ does help improve the retrieval performance and it
also demonstrates the validity of the proposed methodology in
extracting sentenced-based ground truth tag importance. 

\subsection{Performance of Tag Importance Prediction}\label{subsection:TIP}
To evaluate tag importance prediction performance, we first compare the
performance of the state-of-the-art method with that of the proposed tag
importance prediction method on UIUC dataset. Then, under continuous-valued tag importance setting, we study
the effect of different feature types on the proposed tag importance
prediction model, along with the loss introduced by binary-valued importance for all datasets introduced in Section~\ref{dataset}. 

For the purpose of performance benchmarking, we simplified our structured model as discussed in Sec.~\ref{model} to achieve
binary-valued tag importance prediction and compared with
\cite{berg2012understanding}. Here we use accuracy as the evaluation metric. Same as in \cite{berg2012understanding}, accuracy is defined as the percentage of correctly classified object instances against the total number of object instances in test images. For fair comparison, we also ran 10
simulations of 4-fold cross validation, and compared the mean and
standard deviation of estimated accuracy. The performance comparison
results are shown in Table~\ref{tab:benckmarkcvpr2012}.  The baseline
method simply predicts ``yes" (or important) for every object instance
while the next column refers to the best result obtained in the work of
Berg {\em et al.} \cite{berg2012understanding}.  Our simplified
structured model can further improve the prediction accuracy of
\cite{berg2012understanding} by 5.6\%. 
\begin{table}[htb]
\begin{center}
\caption{Performance comparison of tag importance prediction.}\label{tab:benckmarkcvpr2012}
\begin{tabular}{c||c|c|c} \hline
Methods            & Baseline & Berg {\em et al.}\cite{berg2012understanding} & Proposed \\ \hline
Accuracy Mean      & 69.7\%   & 82.0\% &  87.6\%  \\ \hline
Accuracy STD       & 1.3      & 0.9    &  0.7     \\ \hline
\end{tabular}
\end{center}
\end{table}

Next, we evaluate the continuous-valued tag importance prediction performance of 7 different models. Here we use prediction error as the evaluation metric, which is defined as the average MAD in Eq. (\ref{loss}) across all test images. These 7 models are:
\begin{enumerate}
\item Equal importance of all tags (called the Baseline);
\item Visual features only (denoted by Visual);
\item Visual and semantic features (denoted by Visual+Semantic);
\item Visual, semantic and context features (denoted by our Model);
\item Visual, semantic and context features with ground truth bounding boxes (denoted by our Model/True bbox); 
\item Equal importance of tags that are mentioned in any sentence. This corresponds to the true binary-valued tag importance computed as in \cite{berg2012understanding} (denoted by binary true);
\item Equal importance of tags that are predicted as ``important" using the model proposed in \cite{berg2012understanding} (denoted by binary predicted).
\end{enumerate}

For the 2nd and 3rd models, we adopt the ridge regression
models trained by visual features and visual plus semantic features,
respectively. The 4th model is the
proposed structured model as described in Section~\ref{model}. These 3 models help us to understand the impact of different feature types as described in Sec. \ref{cue}. The 5th model differs from the 4th model in that it uses ground truth bounding boxes to compute the visual features. It enables us to identify how object detection error will affect the tag importance prediction. For the 6th and 7th models, they are used to quantify how the binary-valued tag importance (i.e.\ treating important tags as equally important) results in tag importance prediction error, which serves as an indicator of performance loss in retrieval. Specifically, the 6th and 7th models will generate the true and predicted important tags using the method proposed in \cite{berg2012understanding}, respectively. Then, the important tags within the same image will be treated as equally important and assigned the same continuous-valued importance. Note that the 6th model is not achievable but only serves as the best case for the 7th model.

\begin{figure}[!t]
\centering
\subfloat[UIUC dataset.]{\includegraphics[width=0.24\textwidth]{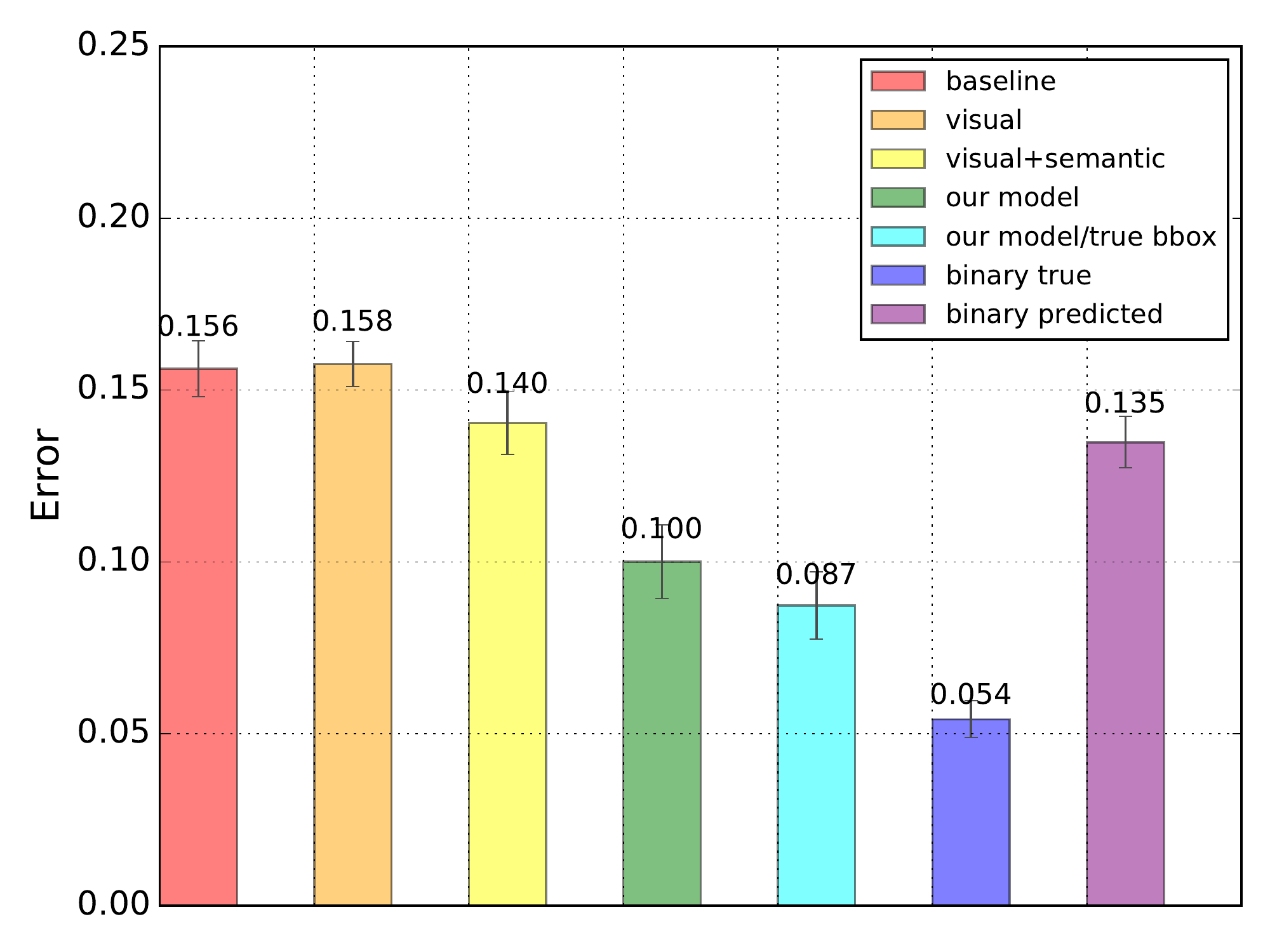}%
\label{fig:UIUC_error}}
\hfil
\subfloat[COCO dataset.]{\includegraphics[width=0.24\textwidth]{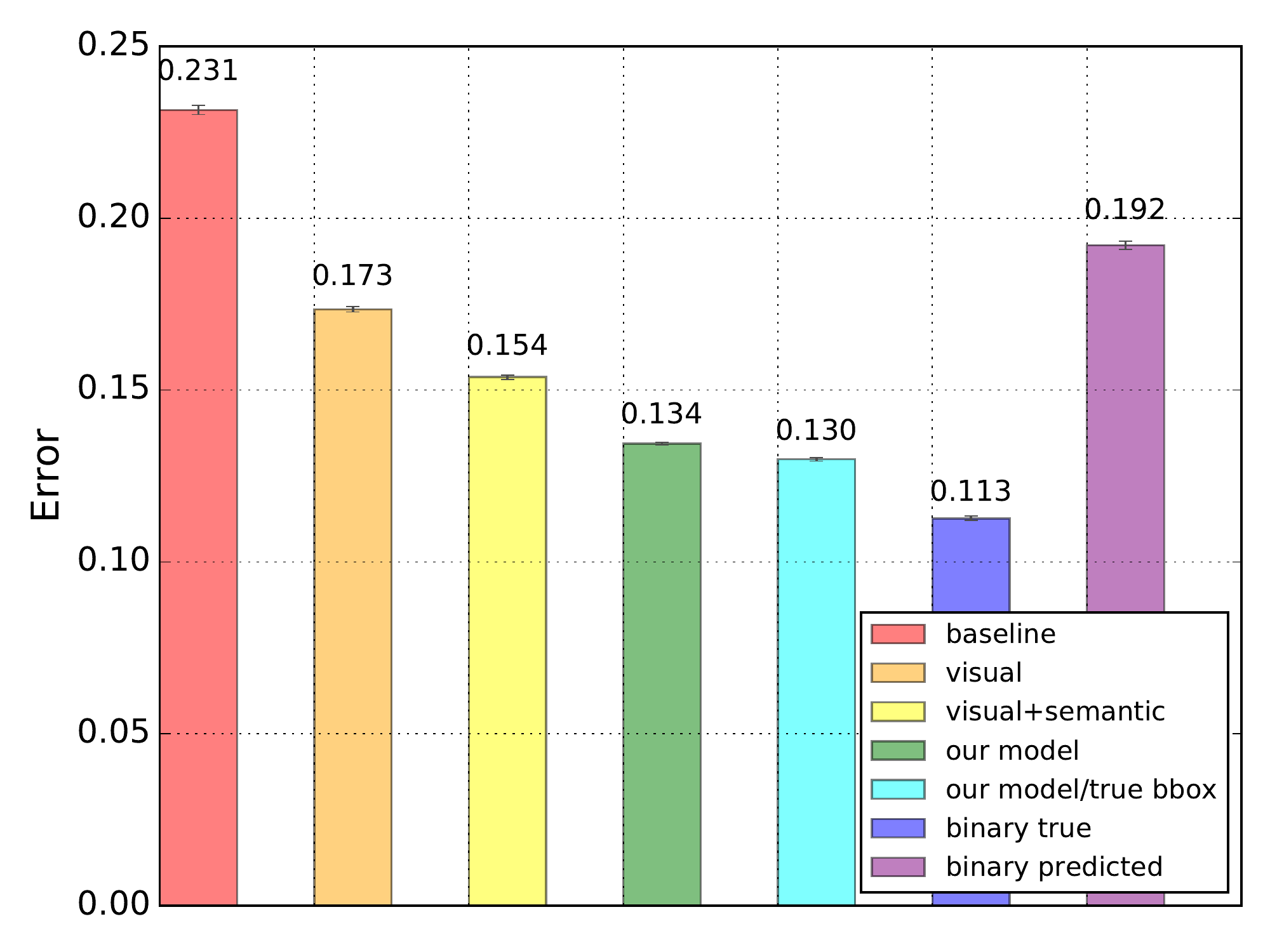}%
\label{fig:COCO_error}}\\
\vspace{-3mm}
\subfloat[COCO Scene dataset.]{\includegraphics[width=0.4\textwidth]{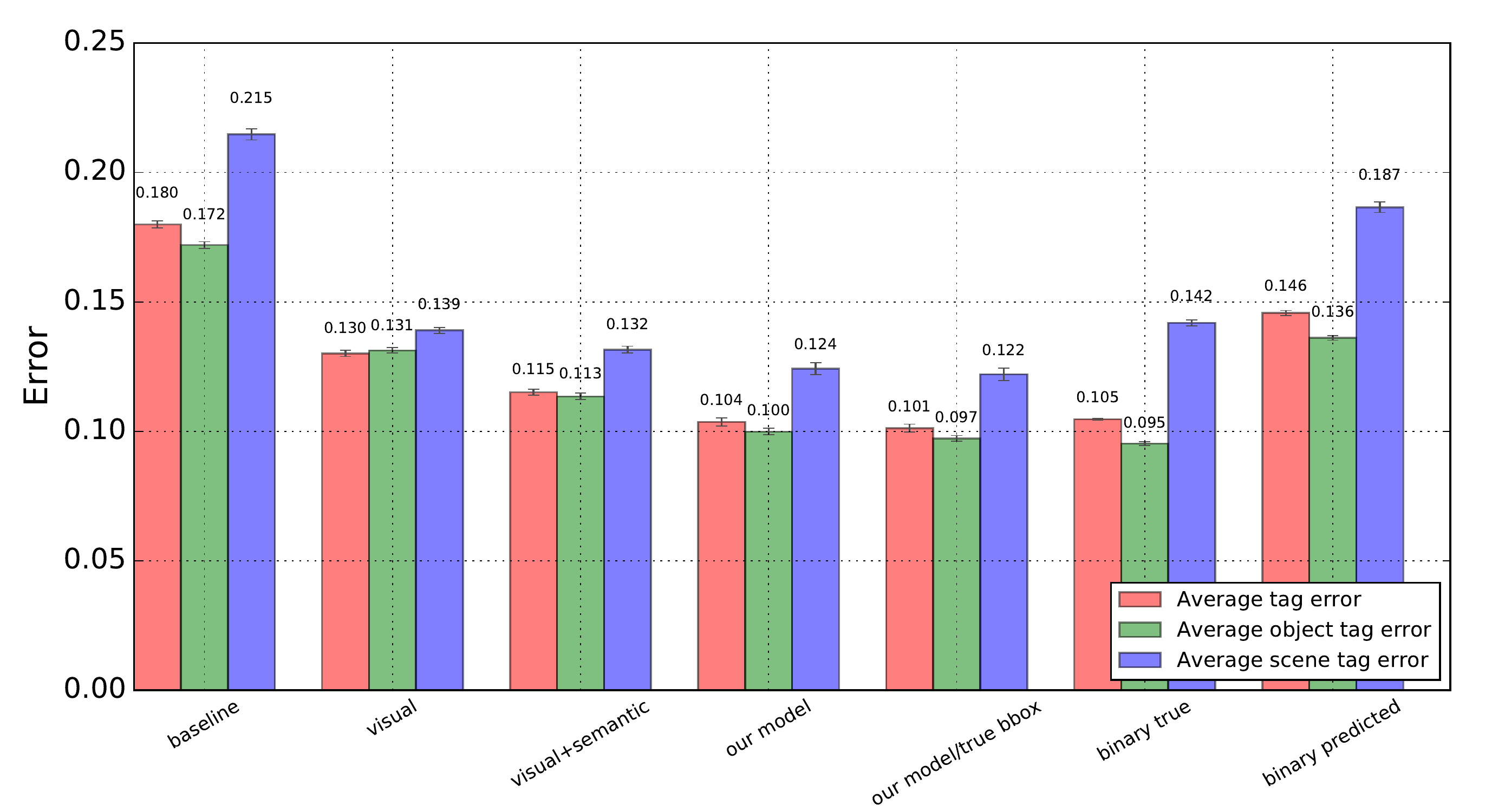}%
\label{fig:COCO_scene_error}}
\caption{Comparison of continuous-valued tag importance prediction errors of seven models: (a) the UIUC dataset, (b) the COCO dataset, and (c) the COCO Scene dataset. }
\label{fig:error}
\end{figure}

We used 5-fold cross validation to evaluate these prediction models.
The prediction errors for UIUC and COCO datasets are
shown in Figs.~\ref{fig:error}(a) and (b), respectively. For COCO Scene dataset, we show the prediction error for all, object, and scene tags in Figs.~\ref{fig:error}(c).

These figures show that our proposed structured prediction model (4th)
can achieve approximately \textbf{40\%} performance gain with respect to the
baseline (1st). For COCO and COCO Scene dataset, we observe performance gain of all 3 feature types, among which visual, semantic, and context features result in approximately 28\%, 11\%, and 11\% prediction error reduction respectively. By comparing the 4th to the 5th model, we find the object detection error only results in approximately 3\% performance loss. Moreover, it is noted that even true binary-valued tag importance lead to non negligible prediction error by ignoring relative importance between tags, and this error will propagate to predicted binary-valued tag importance model, resulting in 48\% and 45\% performance loss over our proposed model on COCO and COCO Scene dataset, respectively. Lastly, for COCO Scene dataset, it is observed that scene tag importance is more
difficult to predict as compared to object tag importance. Thus, the overall error (average over both object and scene tag importance) is higher than the average error of object tag importance but lower than that of scene tag importance.

The tag importance prediction performance on UIUC dataset differs from that of COCO and COCO Scene in two parts: 1) the visual features results in 1\% performance loss compared with baseline; 2) the binary-valued tag importance has less performance loss compared to COCO and COCO Scene. The above phenomena were caused by the bias of the UIUC dataset, in which 454 out of 1000 images have only one tag, and 415 of them have tag with importance value 1. This bias makes modeling the relative importance between tags within the same image insignificant. Thus, the baseline and binary-valued tag importance based models can achieve reasonable performance on UIUC dataset but not on COCO and COCO Scene datasets.

\subsection{Performance of Multimodal Image Retrieval}\label{subsec:MIR}

In this subsection, we show the retrieval experimental
results on COCO and COCO Scene datasets using the settings given in Section~\ref{settings}. For both datasets, we randomly sampled 10\% of images as queries and the other as database images. Among the database images, 50\% were used as MIR training images. For I2I and I2T experiments, we directly used the image as the query. For T2I experiment, weighted tag list based on ground truth tag importance vector was used as query.

\begin{figure}[!t]
\centering
\subfloat[COCO dataset.]{\includegraphics[width=0.24\textwidth]{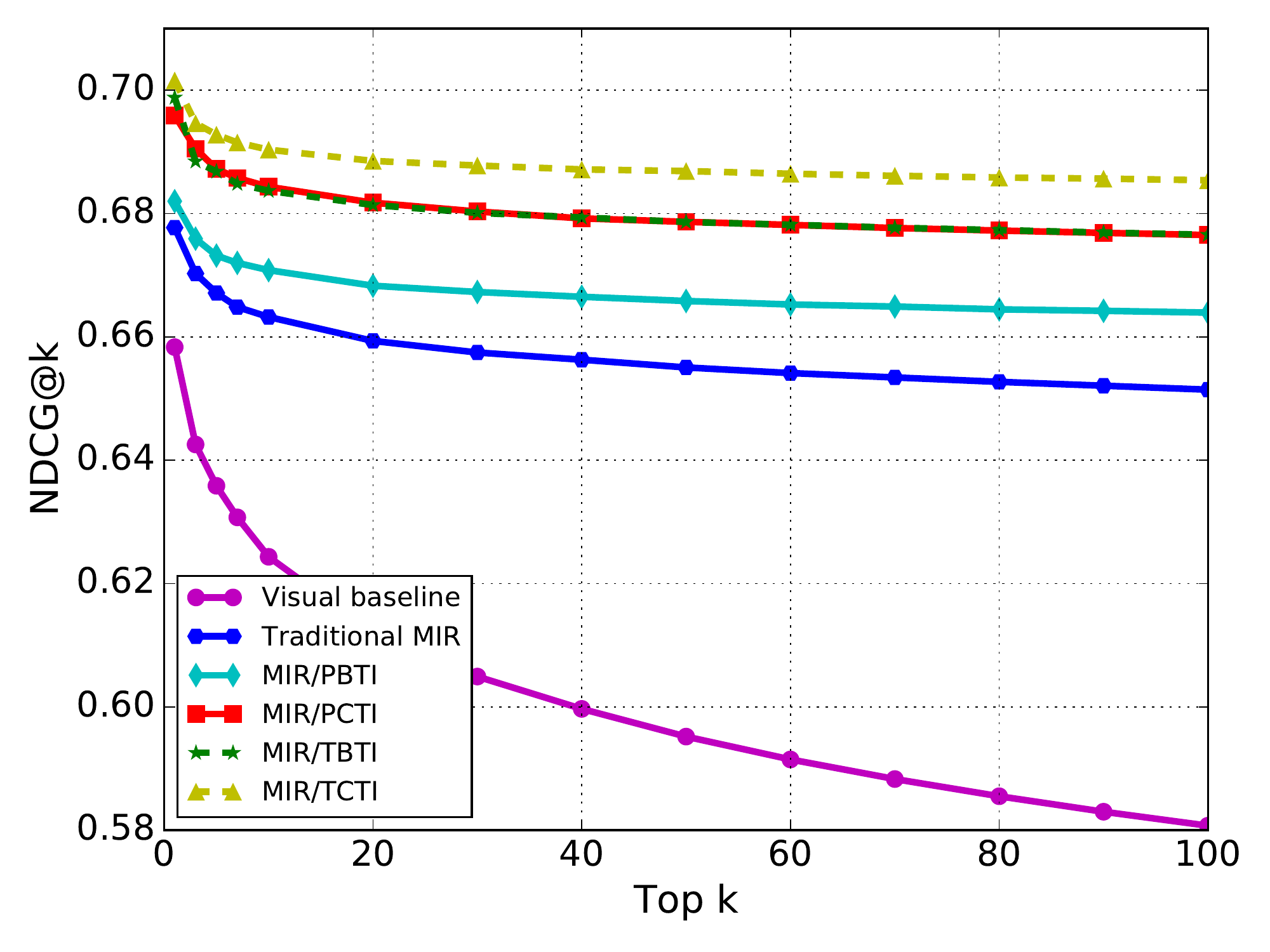}%
\label{fig:ndcg_micro}}
\hfil
\subfloat[COCO Scene dataset.]{\includegraphics[width=0.24\textwidth]{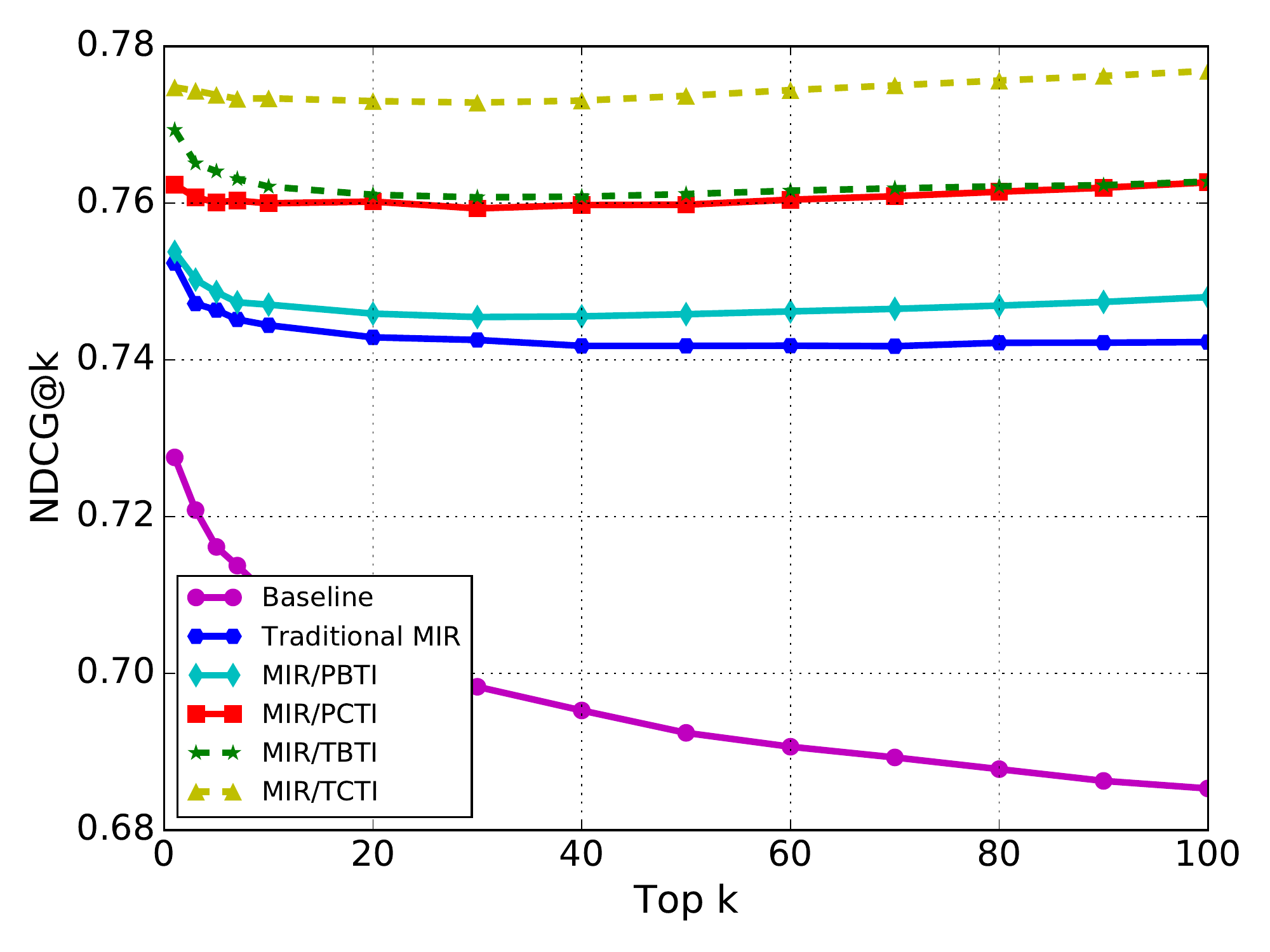}%
\label{fig:ndcg_cocoscene}}
\caption{The NDCG curves for the image-to-image retrieval on the (a) COCO 
and (b) COCO Scene datasets. The dashed lines are upper bounds for importance based MIR systems.}\label{fig:ndcg}
\end{figure}

\begin{figure}[!t]
\centering
{\includegraphics[width=0.5\textwidth]{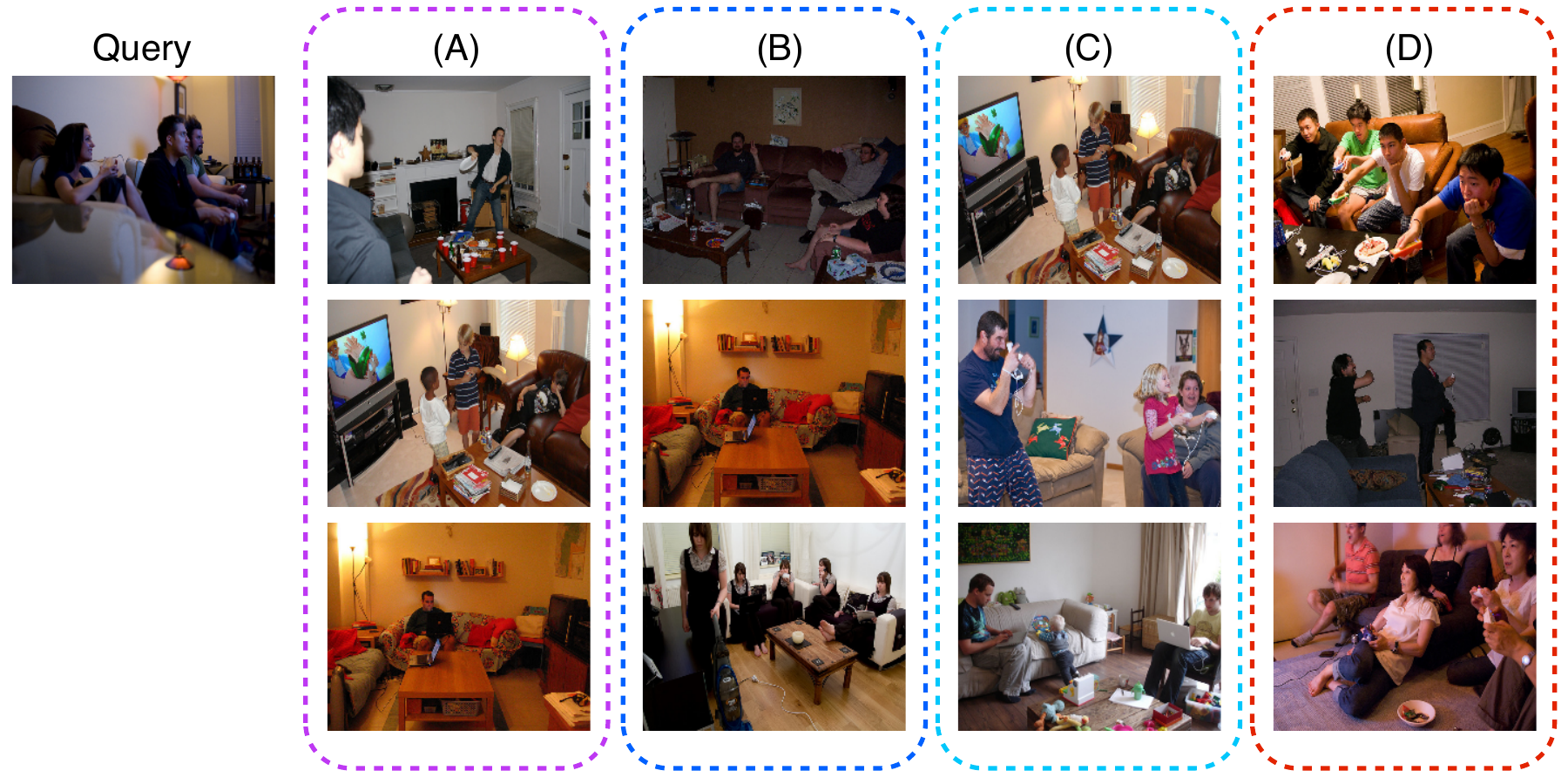}%
\label{fig:coco_scene_object}}
{\includegraphics[width=0.5\textwidth]{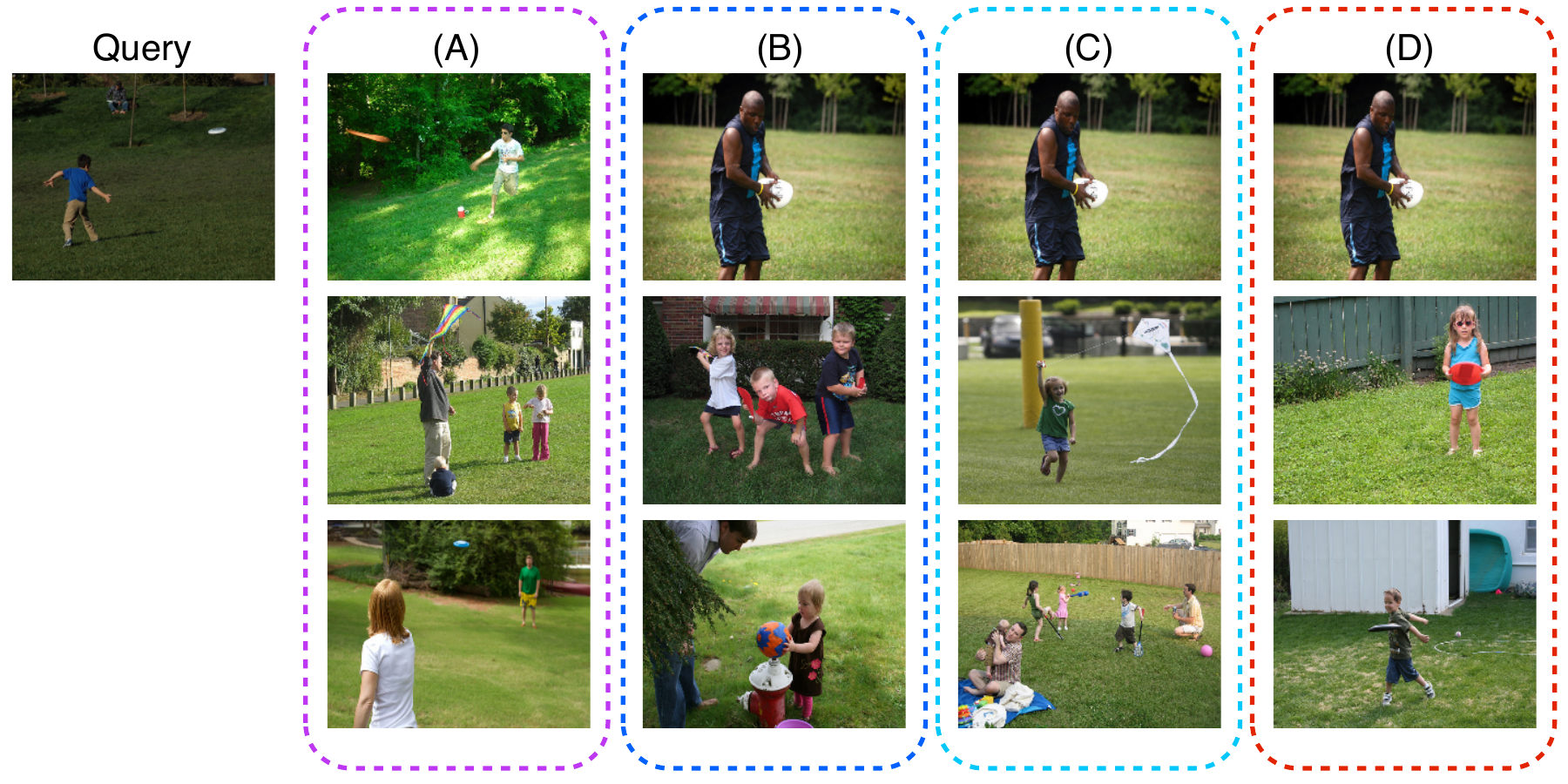}%
\label{fig:coco_scene_scene}}
\caption{Top three I2I retrieved results for two exemplary queries, where the four columns show four retrieval systems: (A)
Visual Baseline, (B) Traditional MIR, (C) MIR/PBTI, and (D)
MIR/PCTI.} \label{fig:coco_scene_visual}
\end{figure}

\textbf{I2I Results.} 
The NDCG curves of COCO and COCO Scene datasets are shown in
\figurename~\ref{fig:ndcg}(a), (b), respectively. We have the following observations from the plots. First, for all datasets, we find significant improvement of all MIR systems over visual baseline. It seems deep features can give reasonable performance on retrieving the most similar image, but its efficiency lags behind MIR systems as $K$ becomes larger. Second, our proposed MIR/PCTI system exhibits considerable improvements over other practical MIR systems, including Traditional MIR and MIR/PBTI. Specifically, for $K=50$ (the typical number of retrieved images user is willing to browse for one query), the MIR/PCTI can achieve approximately 14\% and 10\% gain over visual baseline,  4\% and 2\% over Traditional MIR, and 2\% and  2\% over MIR/PBTI on COCO and COCO Scene datasets, respectively. Moreover, our proposed MIR/PCTI system can even match the upper bound of MIR system using binary-valued importance, and it only has 1\% and 2\% performance gap with its upper bound. Finally, by associating Fig.~\ref{fig:ndcg} to Fig.~\ref{fig:error}, we can identify that the tag importance prediction performance roughly correlates to retrieval performance. Thus, better tag importance prediction leads to better I2I retrieval performance. Some qualitative I2I retrieval results are shown in \figurename~\ref{fig:coco_scene_visual}. Generally speaking, our proposed system can capture the overall semantic of queries more accurately, such as ``person playing wii in the living room" for the 1st query and ``person playing frisbee in yard area" for the 2nd query, while the remaining 3 systems fail to preserve some important objects such as ``remote" or ``frisbee".


\begin{figure}[!t]
\centering
\subfloat[COCO dataset.]{\includegraphics[width=0.24\textwidth]{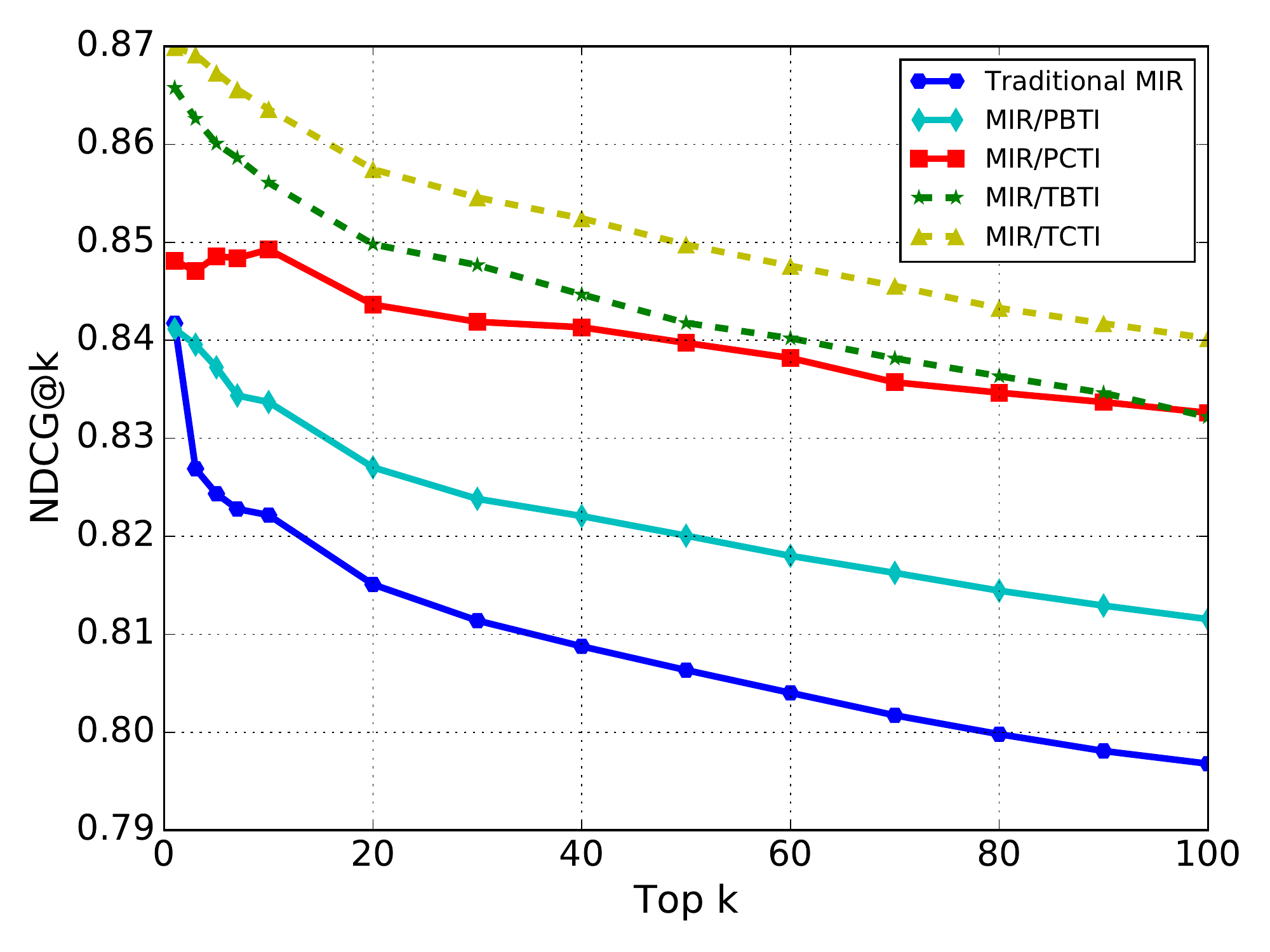}%
\label{fig:ndcg_T2I_micro}}
\hfil
\subfloat[COCO Scene dataset.]{\includegraphics[width=0.24\textwidth]{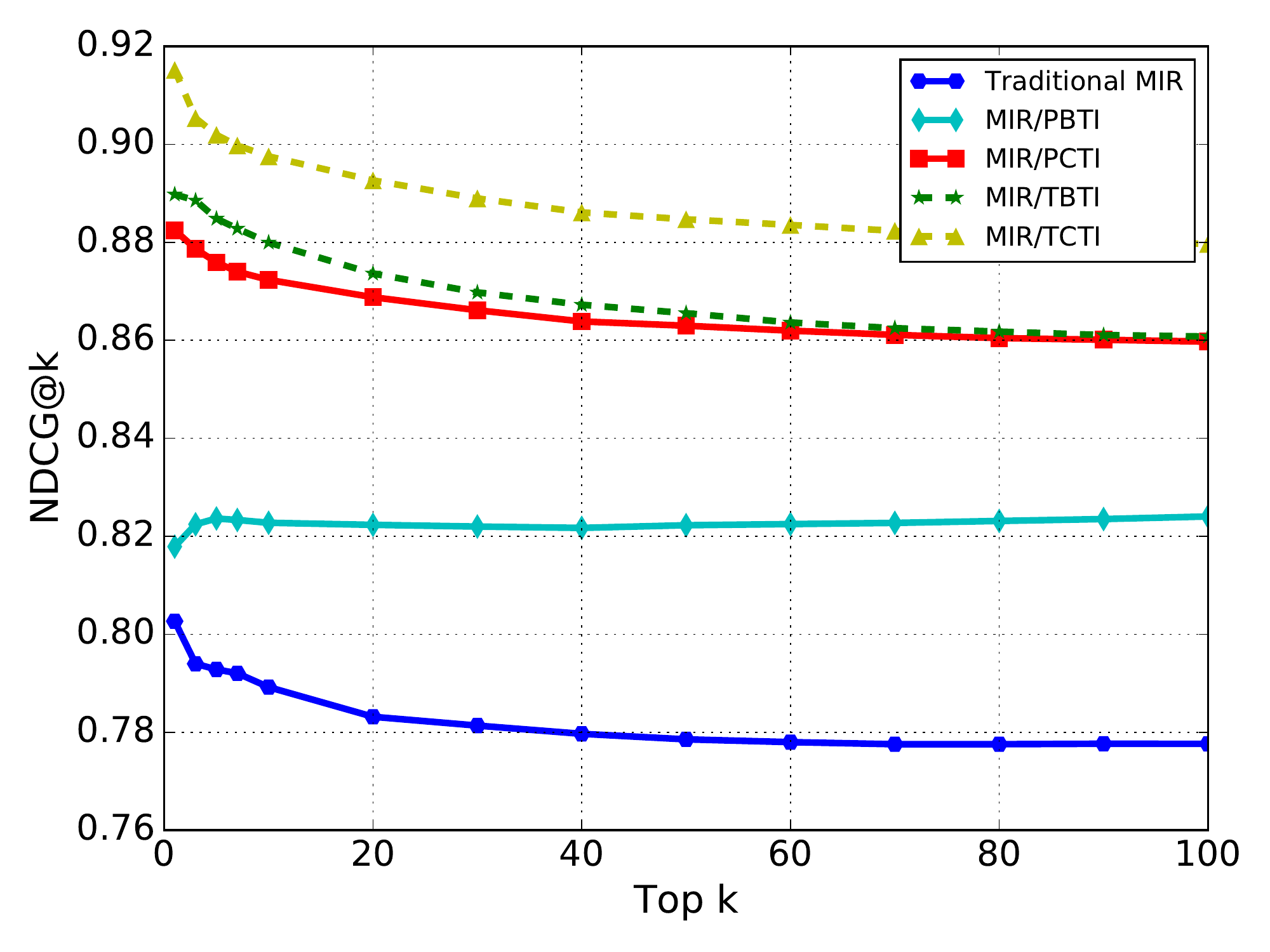}%
\label{fig:ndcg_cocoscene_T2I}}
\caption{The NDCG curves for the tag-to-image retrieval on the (a) COCO 
and (b) COCO Scene datasets. The dashed lines are upper bounds for importance based MIR systems.}\label{fig:ndcg_T2I}
\end{figure}

\begin{figure}[!t]
\centering
\subfloat[A input tag query seeking object centric image.]
{\includegraphics[width=0.45\textwidth]{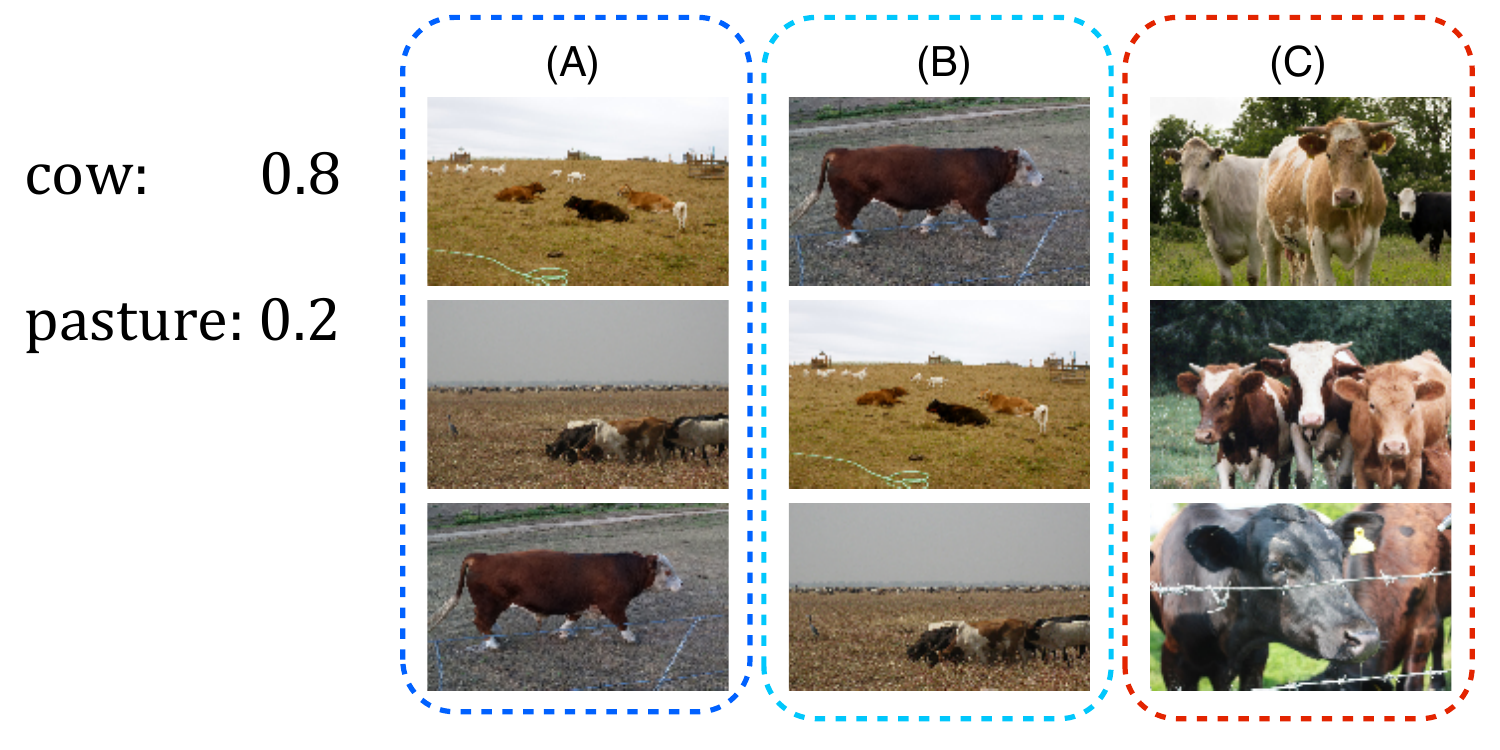}%
\label{fig:T2I1}}\\
\subfloat[A input tag query seeking scene centric image.]
{\includegraphics[width=0.45\textwidth]{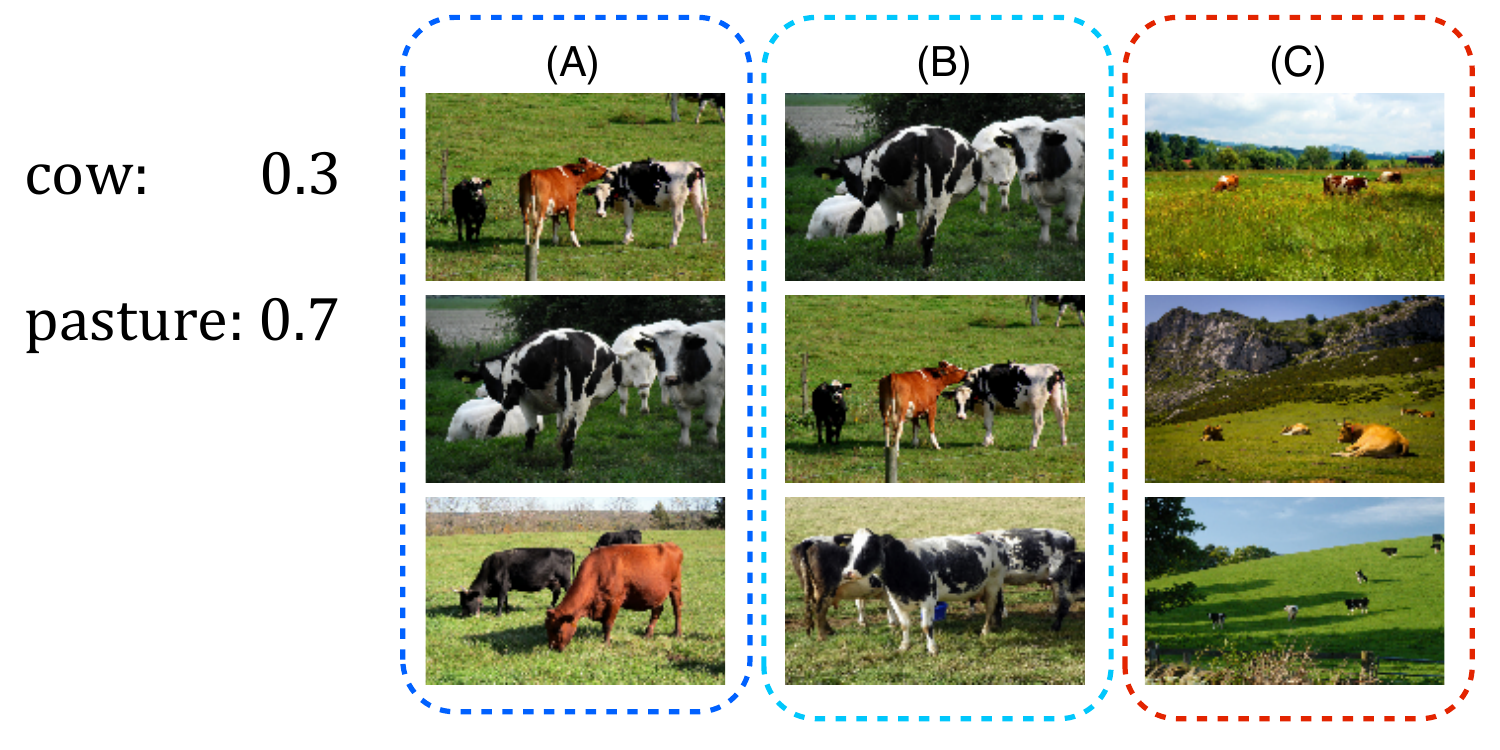}%
\label{fig:T2I2}}
\caption{Tag-to-Image retrieval results for two exemplary query with different focus, where the three columns correspond to the top three ranked tags of three MIR systems: (A)Traditional MIR, (B) MIR/PBTI, and (C) MIR/PCTI. }\label{fig:T2I_sub}
\end{figure}

\textbf{T2I Results.} We show the NDCG curves of T2I results on COCO and COCO Scene datasets in \figurename~\ref{fig:ndcg_T2I} (a) and (b), respectively. Our proposed MIR/PCTI model shows consistent superior performance over Traditional MIR and MIR/PBTI on both datasets. Particularly, for $K$=50, the MIR/PCTI outperforms the Traditional MIR by 4\% and 11\%, and the MIR/PBTI by 2\% and 5\% on COCO and COCO Scene datasets, respectively. Moreover, the proposed MIR/PCTI system only has 1\% and 2\% performance gap with its upper bound on the two datasets.
\figurename~\ref{fig:T2I_sub} shows the two qualitative results of T2I retrieval, where the two input queries consist of the same tag pair but have different focus. It is observed that our proposed system can correctly retrieve scene/object centric images as indicated by the importance value, while the other two systems can not.

\begin{figure}[!t]
\centering
\subfloat[COCO dataset.]{\includegraphics[width=0.24\textwidth]{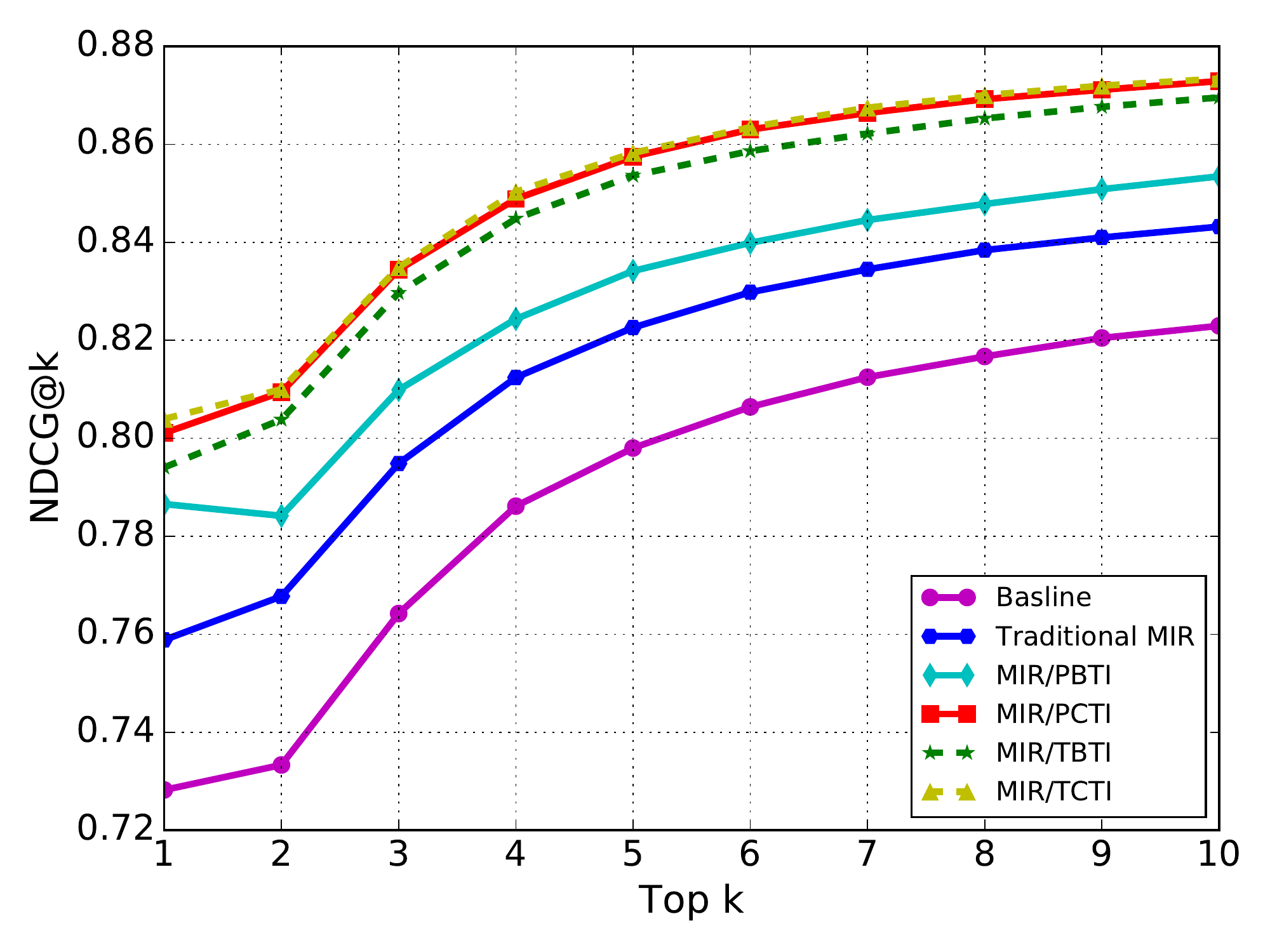}%
\label{fig:ndcg_T2I_micro}}
\subfloat[COCO Scene dataset.]{\includegraphics[width=0.24\textwidth]{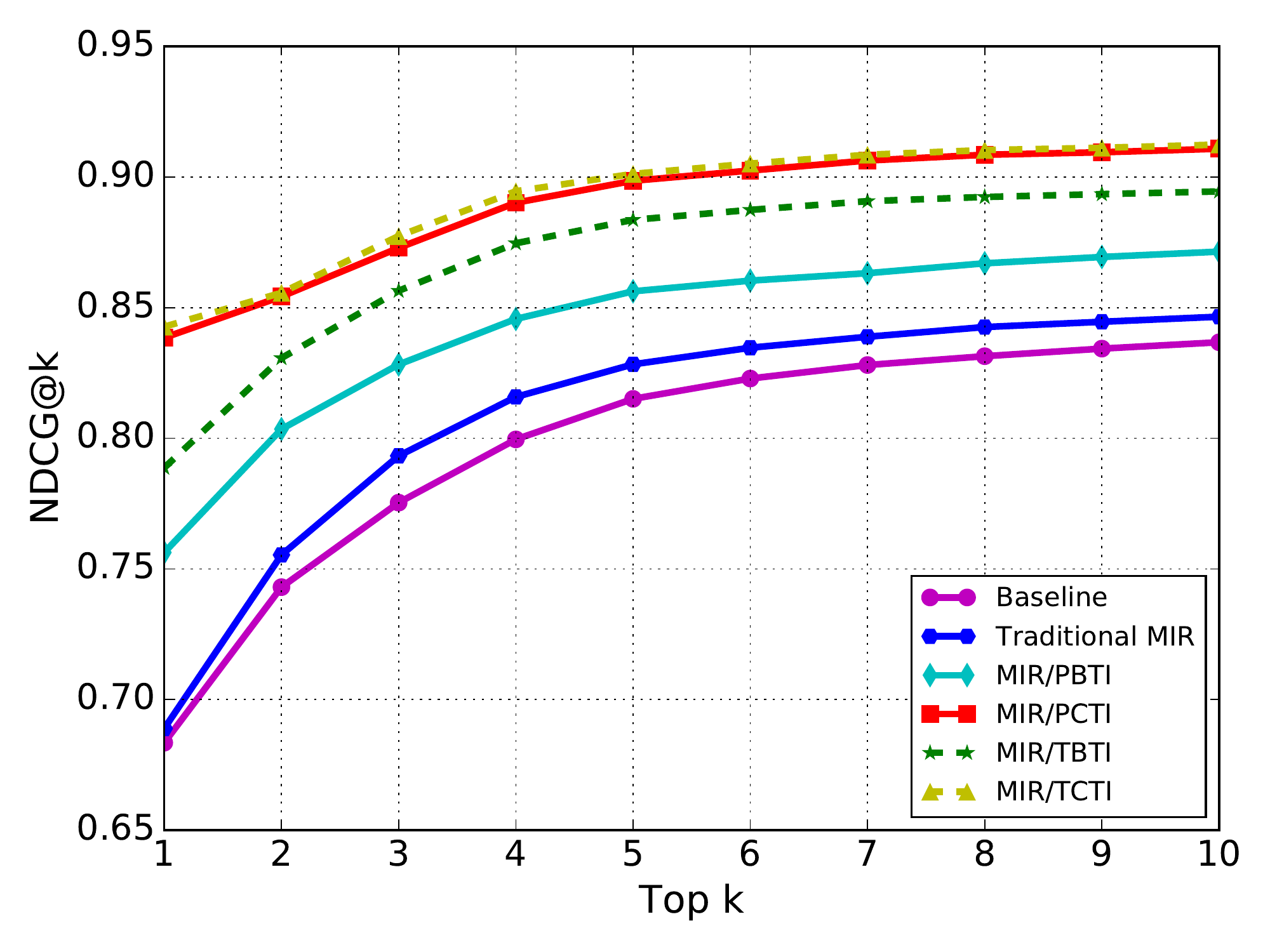}%
\label{fig:ndcg_cocoscene_I2T}}
\caption{The NDCG curves for auto ranked tag list generation on the (a) COCO 
and (b) COCO Scene datasets. The dashed lines are upper bounds for importance based MIR systems.}\label{fig:ndcg_I2T}
\end{figure}

\begin{figure}[!t]
\centering
{\includegraphics[width=0.45\textwidth]{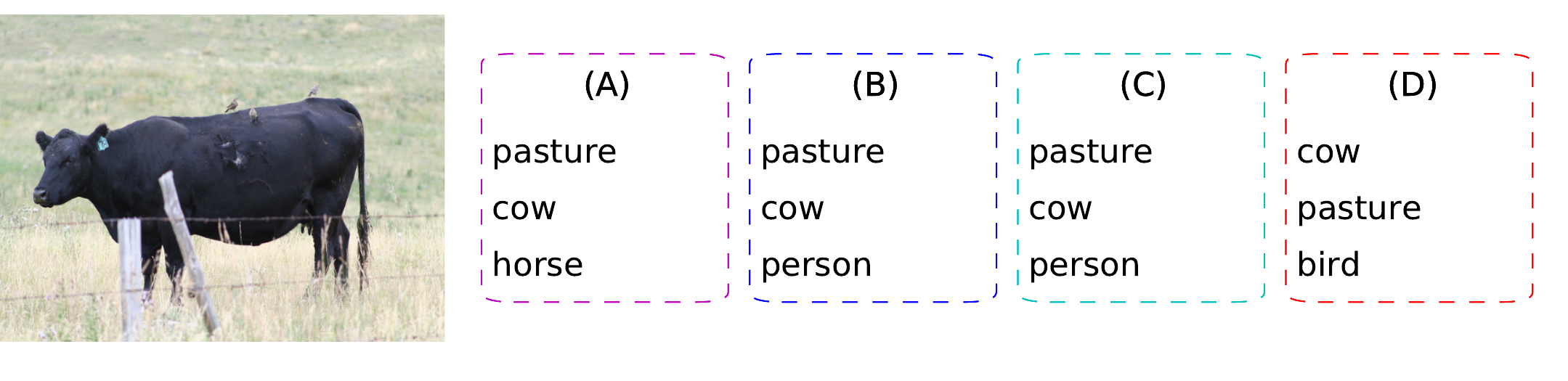}%
\label{fig:I2T1}}
\hfil
{\includegraphics[width=0.45\textwidth]{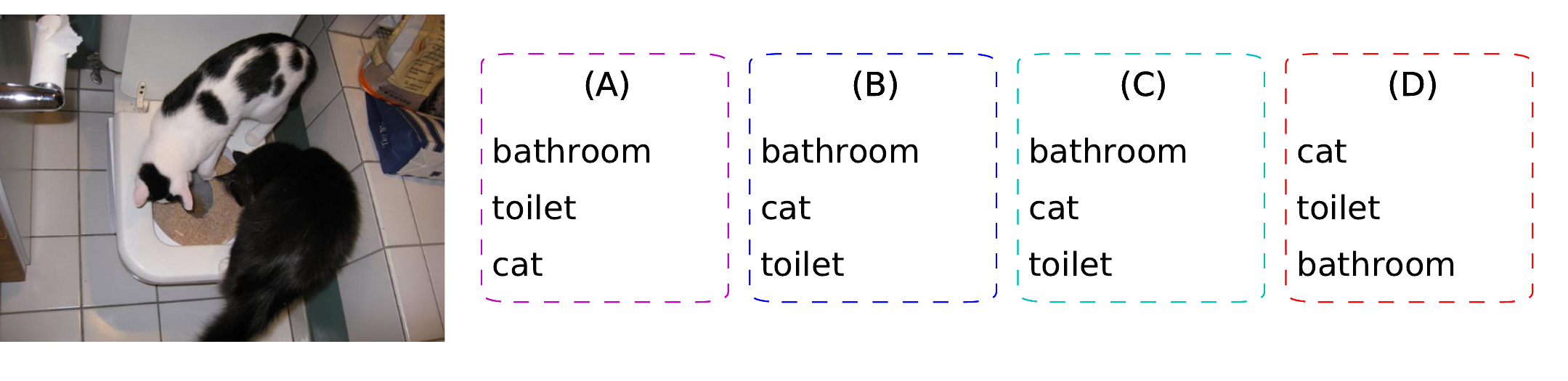}%
\label{fig:I2T2}}
\vspace{-5mm}
\caption{Tagging results for two exemplary images, where the four columns correspond to the top three ranked tags of four MIR systems: (A) Baseline, (B) Traditional MIR, (C) MIR/PBTI, and (D) MIR/PCTI. }\label{fig:I2T_sub}
\end{figure}

\textbf{I2T Results.} The results of tagging on COCO and COCO Scene datasets are shown in \figurename~\ref{fig:ndcg_I2T} (a) and (b), respectively. Again, we observe consistent improvements of MIR/PCTI over Traditional MIR and MIR/PBTI. Specifically, for $K=3$ (the typical number of tags each image have in these two datasets), the MIR/PCTI can achieve approximately 8\% and 13\% gain over baseline,  5\% and 10\% over Traditional MIR, and 3\% and  5\% over MIR/PBTI on COCO and COCO Scene datasets, respectively. More surprisingly, its performance can outperform the upper bound of MIR/PBTI and match that of MIR/TCTI. This suggests that our proposed system can not only generate tags but also rank them according to their importance.
Sample qualitative tagging results are shown in  \figurename~\ref{fig:I2T_sub}, in which we can see that our proposed model will rank more important tags (``cow" and ``cat") ahead of unimportant/wrong ones (``person" and ``bathroom").

\section{Conclusion and Future Work}\label{conclusion}

A multimodal image retrieval scheme based on tag importance prediction
(MIR/TIP) was proposed in this work.  Both object and scene tag
importance were measured from human sentence descriptions and used as the
ground truth based on a Natural Language Processing methodology.  Three
types of features (namely, semantic, visual and context) were identified
and a structured model was proposed for object and scene tag importance
prediction. Model parameters were trained using the Structural Support
Vector Machine formulation.  It was shown by experimental results that
the proposed sentence-based tag importance measure and the proposed tag
importance prediction can significantly boost the performance of various retrieval tasks. 

To make the system more practical to real world applications, it is worthwhile to extend the importance idea to other types of tags.
Specifically, it would be interesting to extend the idea to scene graph based Visual Genome dataset \cite{krishnavisualgenome} with more densely annotated object bounding boxes, attributes and relationship information. Furthermore, 
it is promising to improve retrieval performance by training Convolutional Neural Network and Canonical Correlation Analysis jointly using end-to-end learning. Finally, how to jointly optimize tag importance prediction and retrieval system remains an open question and will be explored in our future work.

\ifCLASSOPTIONcaptionsoff
  \newpage
\fi
\bibliographystyle{IEEEtran}
\bibliography{reference}

%
%
%
\vspace{-13mm}
\begin{IEEEbiography}[{\includegraphics[width=1in,height=1.25in,clip,keepaspectratio]{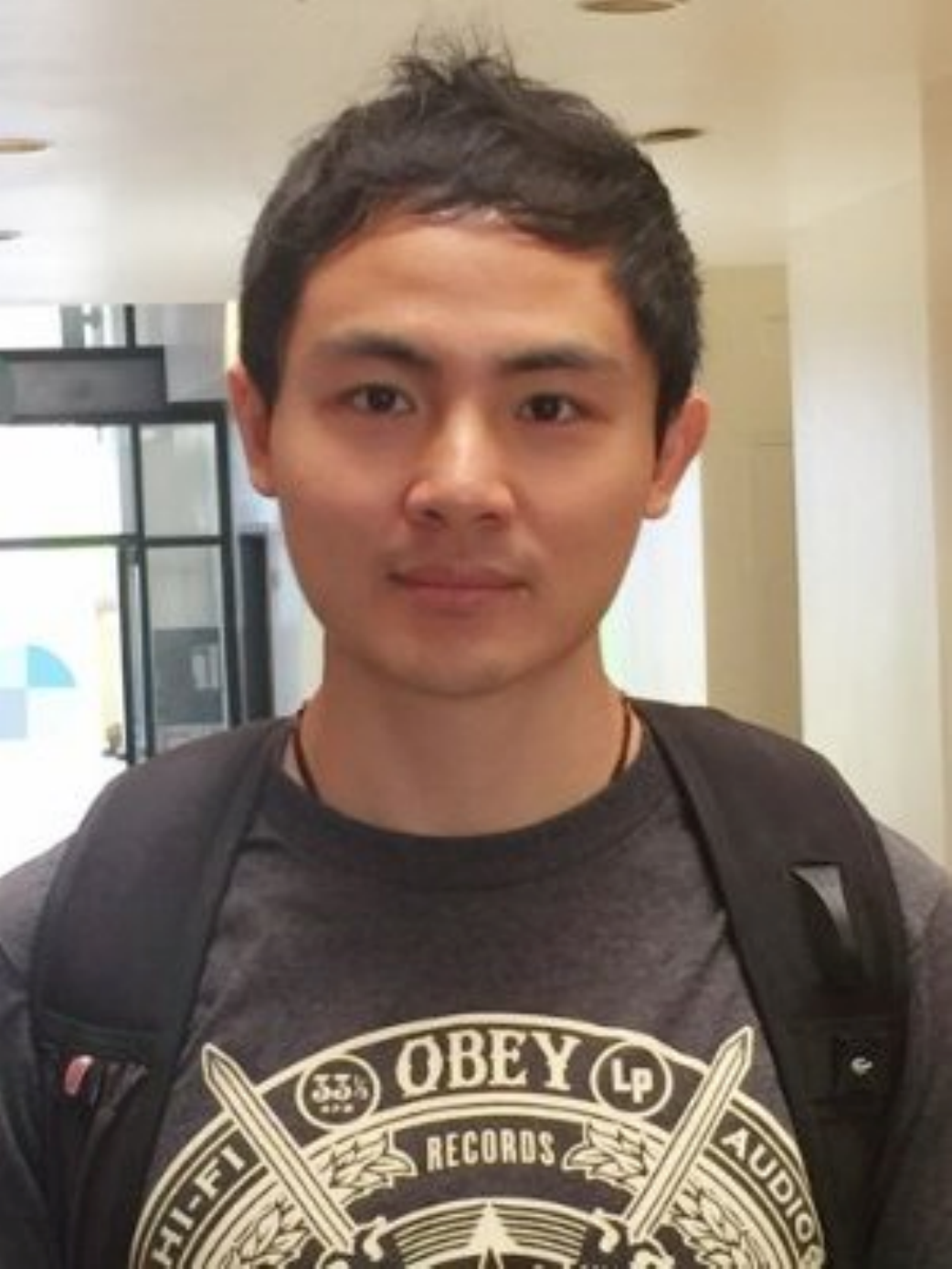}}]{Shangwen Li}
received the B.S. and M.S. degrees, both in electrical engineering, from the Zhejiang University, Hangzhou, China in 2008 and 2011 respectively. Since August 2012, he has been been with Media Communications Lab at University of Southern California (USC), Los Angeles. His research interests include image retrieval, scene understanding and object detection using machine learning techniques.
\end{IEEEbiography}
\vspace{-13mm}
\begin{IEEEbiography}[{\includegraphics[width=1in,height=1.25in,clip,keepaspectratio]{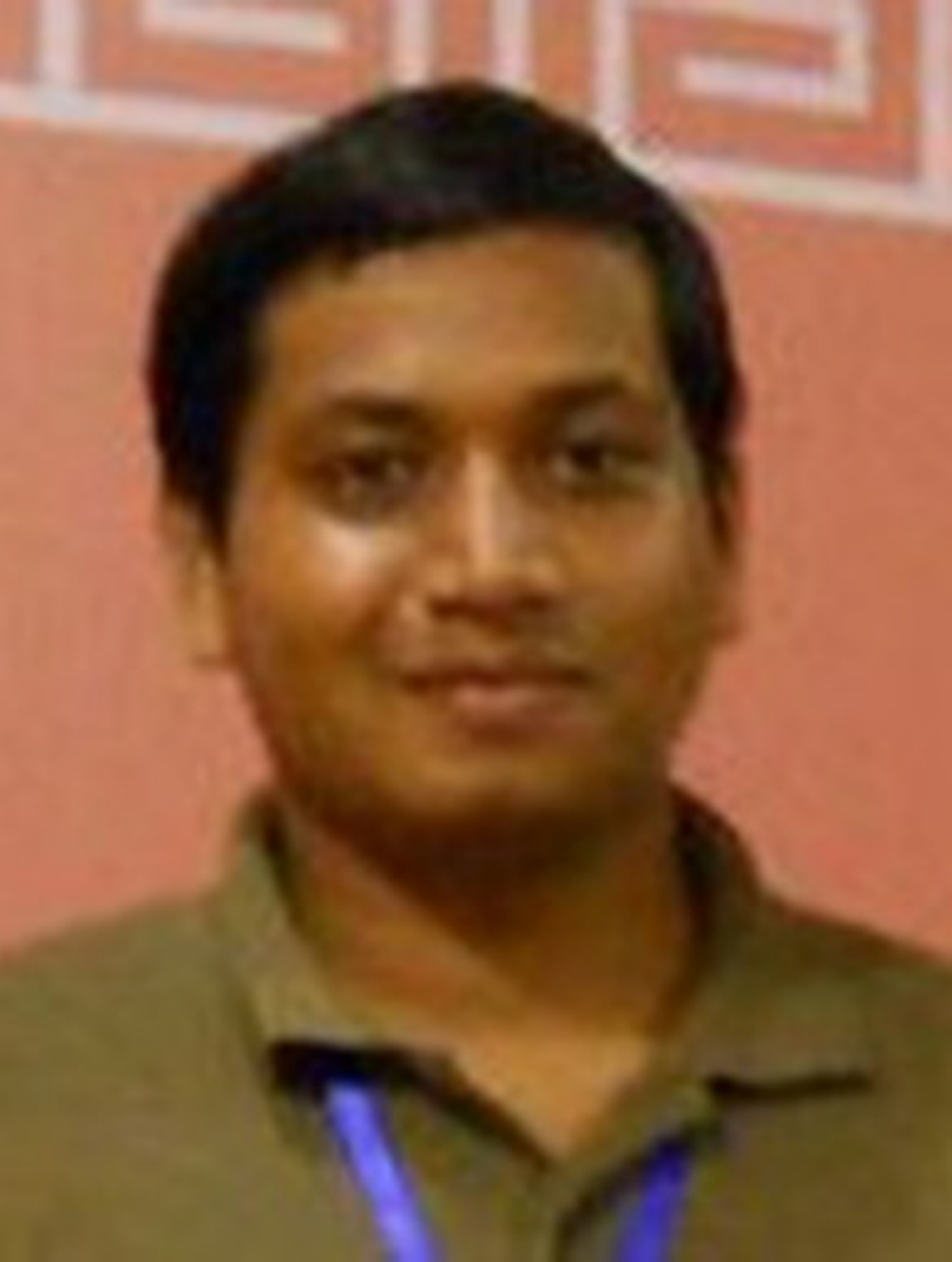}}]{Sanjay Purushotham}
received his PhD in Electrical Engineering at the University of Southern California (USC) in 2015, where he was a research assistant in the Media Communications Lab (MCL) advised by Prof. C.-C. Jay Kuo. In Sept. 2015, he joined the Department of Computer Science at USC as a Post-doc scholar. His research interests are in machine learning and computer vision.
\end{IEEEbiography}
\vspace{-13mm}
\begin{IEEEbiography}[{\includegraphics[width=1in,height=1.25in,clip,keepaspectratio]{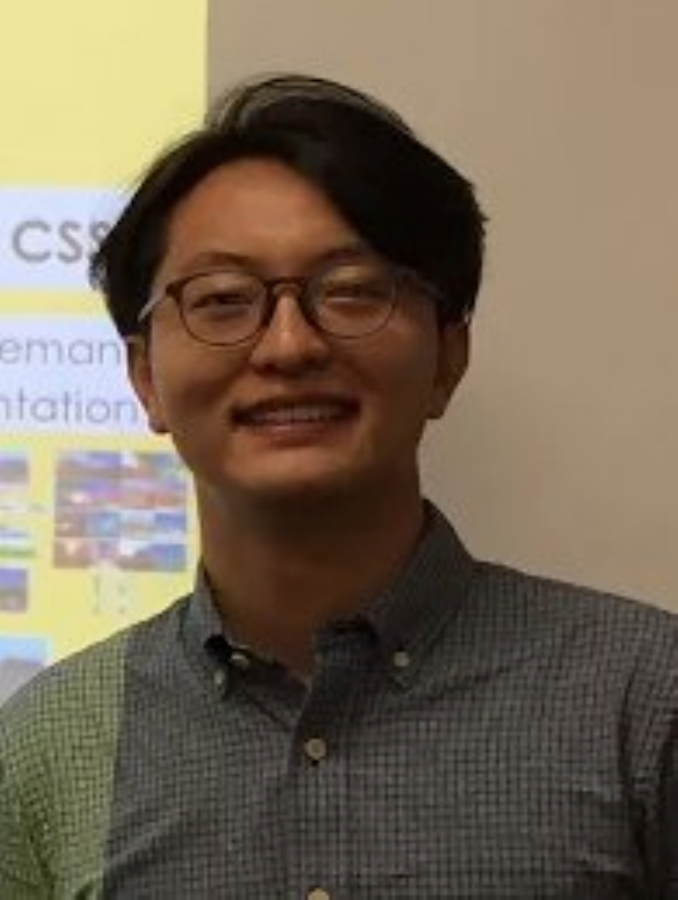}}]{Chen Chen}
received his EE bachelor degree from the Beijing University of Posts and Telecoms (BUPT) in 2010. He is now working on computer vision researches as a PhD candidate in the Media and Creative Lab of the Viterbi Engineer School in USC. His research focuses on scene image segmentation, classification and retrieval using advanced machine learning technologies.
\end{IEEEbiography}
\vspace{-13mm}
\begin{IEEEbiography}[{\includegraphics[width=1in,height=1.25in,clip,keepaspectratio]{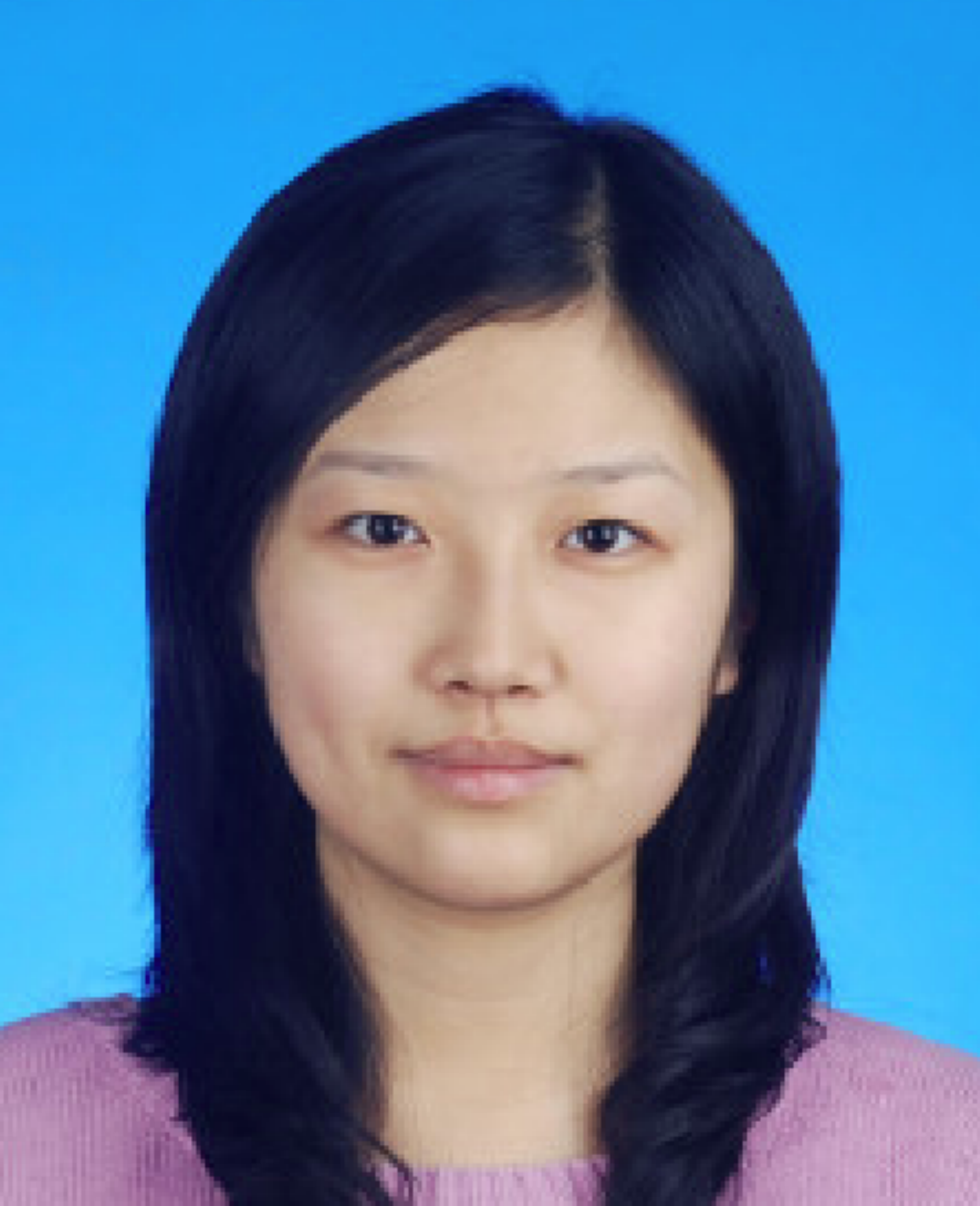}}]{Yuzhuo Ren}
received her B.S. degree from Hebei University of Technology, China, in 2011 and the M.S. degree from University of Southern California, USA, in 2013, both in electrical engineering. She is currently a Ph.D student in USC Media Communications Lab supervised by Prof. C.-C. Jay Kuo. Her research interest is the field of scene understanding, image segmentation, 3D layout estimation using computer vision and machine learning techniques.
\end{IEEEbiography}
\vspace{-13mm}
\begin{IEEEbiography}[{\includegraphics[width=1in,height=1.25in,clip,keepaspectratio]{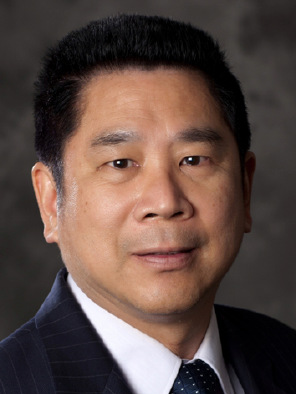}}]{C.-C. Jay Kuo}
received the B.S. degree from National Taiwan University, Taipei, Taiwan, in 1980, and the
M.S. and Ph.D. degrees from the Massachusetts Institute of Technology, Cambridge, MA, USA, in 1985
and 1987, respectively, all in electrical engineering. He is currently the Director of the Multimedia
Communications Laboratory and DeanÕs Professor of Electrical Engineering at the University of Southern
California, Los Angeles, CA, USA. His research interests include digital image/video analysis and modeling,
multimedia data compression, communication and networking, computer vision and machine learning. He
has co-authored about 230 journal papers, 870 conference papers, and 13 books. He is a fellow of the
Institute of Electrical and Electronics Engineers (IEEE), the American Association for the Advancement of
Science (AAAS) and the International Society for Optical Engineers (SPIE).
\end{IEEEbiography}
%

%
%
%
\end{document}